\documentclass[11pt, a4paper, logo, onecolumn, copyright]{googledeepmind}

\usepackage[authoryear, round]{natbib}
\usepackage{cleveref}

\usepackage{amsmath}
\usepackage{amsfonts}
\usepackage{amssymb}
\usepackage{graphicx}
\usepackage{placeins}

\usepackage{subcaption}
\usepackage{float}
\usepackage{fancyvrb,fvextra}
\fvset{listparameters=\setlength{\topsep}{0pt}\setlength{\partopsep}{0pt}}
\usepackage{listings}
\usepackage{tikz}
\usetikzlibrary{shapes,arrows}
\usetikzlibrary{patterns}
\usetikzlibrary{decorations.pathreplacing, decorations.pathmorphing, calligraphy}
\tikzstyle{line} = [draw, very thick, -latex']
\tikzstyle{arrow} = [draw, ->, thick]

\definecolor{bpurp}{HTML}{984ea3}
\definecolor{bblue}{HTML}{377eb8}
\definecolor{bgreen}{HTML}{4daf4a}
\definecolor{borange}{HTML}{ff7f00}
\definecolor{bred}{HTML}{a50f15}
\definecolor{arbitrary}{HTML}{f1b6da}
\definecolor{consistent}{HTML}{80cdc1}
\definecolor{nonsense}{HTML}{737373} 
\definecolor{violate}{HTML}{8e1252}

\title{Scaling Instructable Agents Across Many Simulated Worlds}


\renewcommand{\today}{}

\author[ ]{SIMA Team:\textsuperscript{1}}
\author[ ]{Maria Abi Raad, Arun Ahuja, Catarina Barros, Frederic Besse, Andrew Bolt, Adrian Bolton, Bethanie Brownfield, Gavin Buttimore, Max Cant, Sarah Chakera, Stephanie C. Y. Chan, Jeff Clune\textsuperscript{1,3}, Adrian Collister, Vikki Copeman\textsuperscript{2}, Alex Cullum, Ishita Dasgupta, Dario de Cesare, Julia Di Trapani, Yani Donchev, Emma Dunleavy, Martin Engelcke, Ryan Faulkner, Frankie Garcia, Charles Gbadamosi, Zhitao Gong, Lucy Gonzales\textsuperscript{2}, Karol Gregor, Kshitij Gupta\textsuperscript{2}, Arne Olav Hallingstad, Tim Harley, Sam Haves, Felix Hill, Ed Hirst, Drew A. Hudson, Jony Hudson, Steph Hughes-Fitt, Danilo J. Rezende, Mimi Jasarevic, Laura Kampis, Rosemary Ke, Thomas Keck, Junkyung Kim, Oscar Knagg, Kavya Kopparapu, Rory Lawton, Andrew Lampinen, Shane Legg, Alexander Lerchner, Marjorie Limont, Yulan Liu, Maria Loks-Thompson, Joseph Marino, Kathryn Martin Cussons\textsuperscript{2}, Loic Matthey, Siobhan Mcloughlin, Piermaria Mendolicchio, Hamza Merzic, Anna Mitenkova, Alexandre Moufarek, Valeria Oliveira, Yanko Oliveira, Hannah Openshaw, Renke Pan, Aneesh Pappu, Alex Platonov, Ollie Purkiss, David Reichert, John Reid, Pierre Harvey Richemond, Tyson Roberts, Giles Ruscoe, Jaume Sanchez Elias, Tasha Sandars\textsuperscript{2}, Daniel P. Sawyer, Tim Scholtes, Guy Simmons, Daniel Slater, Hubert Soyer, Heiko Strathmann, Peter Stys, Allison C. Tam\textsuperscript{2}, Denis Teplyashin, Tayfun Terzi, Davide Vercelli, Bojan Vujatovic, Marcus Wainwright, Jane X. Wang, Zhengdong Wang, Daan Wierstra\textsuperscript{2}, Duncan Williams, Nathaniel Wong, Sarah York, Nick Young}
\affil[1]{Google DeepMind unless otherwise noted, authors listed in alphabetical order, contributions listed at end of report}
\affil[2]{work performed while at Google DeepMind}
\affil[3]{University of British Columbia}

\begin{abstract}
Building embodied AI systems that can follow arbitrary language instructions in any 3D environment is a key challenge for creating general AI. Accomplishing this goal requires learning to ground language in perception and embodied actions, in order to accomplish complex tasks. The Scalable, Instructable, Multiworld Agent (SIMA) project tackles this by training agents to follow free-form instructions across a diverse range of virtual 3D environments, including curated research environments as well as open-ended, commercial video games. Our goal is to develop an instructable agent that can accomplish anything a human can do in any simulated 3D environment. Our approach focuses on language-driven generality while imposing minimal assumptions. Our agents interact with environments in real-time using a generic, human-like interface: the inputs are image observations and language instructions and the outputs are keyboard-and-mouse actions. This general approach is challenging, but it allows agents to ground language across many visually complex and semantically rich environments while also allowing us to readily run agents in new environments. In this paper we describe our motivation and goal, the initial progress we have made, and promising preliminary results on several diverse research environments and a variety of commercial video games.
\end{abstract}

\begin{document}

\maketitle

\section{Introduction}
\label{sec:intro}

Despite the impressive capabilities of large language models \citep{brown2020language,hoffmann2022training,openai2023gpt4,anil2023palm,geminiteam2023gemini}, connecting them to the embodied world that we inhabit remains challenging. Modern AI can write computer programs \citep{li2022competition} or play chess at super-human level \citep{silver2018general}, but the ability of AI to perceive and act in the world remains far below human level. Competence in language alone is easier for AI than grounded perception and behavior, underscoring the well-known paradox that what is easier for AI is harder for humans, and vice versa \citep{moravec1988mind}.

Yet, language is most useful in the abstractions it conveys about the world. Language abstractions can enable efficient learning and generalization \citep{hill2020grounded,colas2020language,lampinen2022tell,tam2022semantic,hu2023thought}. Once learned, language can unlock planning, reasoning \citep[e.g.,][]{huang2022language,brohan2023can,driess2023palm,kim2023language}, and communication \citep{zeng2022socratic} about grounded situations and tasks. In turn, grounding language in rich environments can make a system's understanding of the language itself more systematic and generalizable \citep{hill2019environmental}.
Thus, several questions emerge: How can we bridge the divide between the symbols of language and their external referents \citep[cf.,][]{harnad1990symbol}? How can we connect the abstractions and generality afforded by language to grounded perception and action, and how can we do so in a safe and scalable way?

Here, we draw inspiration from these questions---and the prior and concurrent research projects that have addressed them \citep[e.g.,][]{hermann2017grounded,abramson2020imitating,brohan2023rt2,brohan2023can,driess2023palm,wang2023jarvis,tan2024towards}---to attempt to connect language to grounded behavior at scale. Bridging this gap is a core challenge for developing general \emph{embodied AI}.
 
\begin{figure*}
    \centering
    \includegraphics[width=\linewidth]{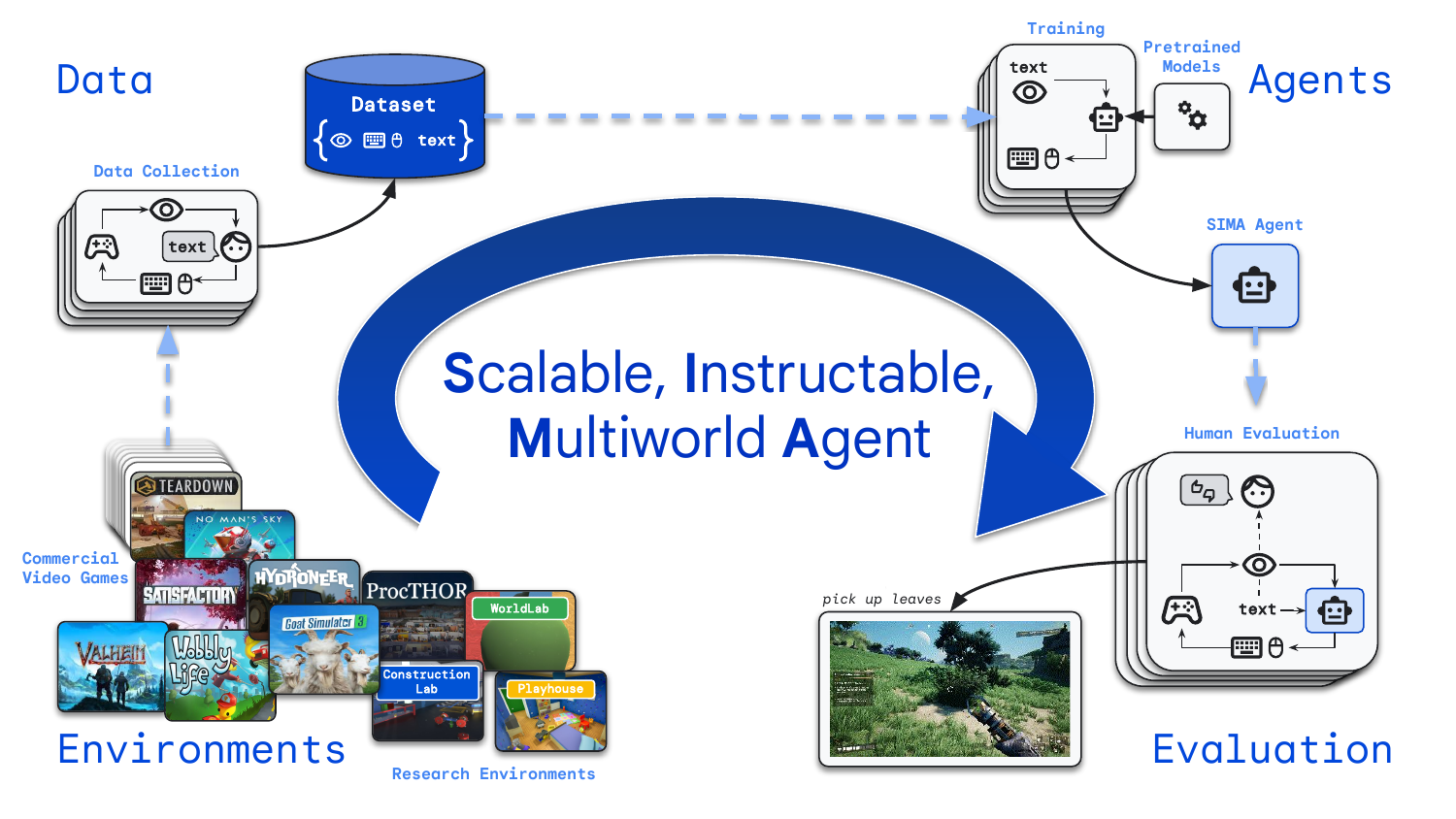}
    \caption{\textbf{Overview of SIMA.} In SIMA, we collect a large and diverse dataset of gameplay from both curated research environments and commercial video games. This dataset is used to train agents to follow open-ended language instructions via pixel inputs and keyboard-and-mouse action outputs. Agents are then evaluated in terms of their behavior across a broad range of skills.}
    \label{fig:overview}
\end{figure*}

The Scalable, Instructable, Multiworld Agent (SIMA) project aims to build a system that can follow \emph{arbitrary} language instructions to act in \emph{any} virtual 3D environment via keyboard-and-mouse actions---from custom-built research environments to a broad range of commercial video games.
There is a long history of research in creating agents that can interact with video games or simulated 3D environments \citep[e.g.,][]{mnih2015human,berner2019dota,vinyals2019grandmaster,baker2022video} and even follow language instructions in a limited range of environments \citep[e.g.,][]{abramson2020imitating,lifshitz2023steve}.
In SIMA, however, we are drawing inspiration from the lesson of large language models that training on a broad distribution of data is the most effective way to make progress in general AI \citep[e.g.,][]{brown2020language,hoffmann2022training,openai2023gpt4,anil2023palm,geminiteam2023gemini}.
Thus, in contrast to prior works \citep[e.g.,][]{abramson2020imitating,vinyals2019grandmaster,berner2019dota,lifshitz2023steve}, we are attempting to tackle this problem across many simulated environments, in the most general and scalable way possible, by making few assumptions beyond interacting with the environments in the same way as humans do.

To this end, have made a number of design decisions that make our approach more general, but also more challenging:
\begin{itemize}
\item We incorporate many rich, visually complex, open-ended video games containing hundreds of objects in a scene and a large number of possible interactions.
\item These environments are asynchronous \citep[e.g.,][]{berner2019dota,vinyals2019grandmaster}; unlike many research environments, they do not stop and wait while the agent computes its next action.
\item Each instance of a commercial video game needs to run on a GPU; thus, we cannot run hundreds or thousands of actors per game per experiment as often done in RL \citep[cf.,][]{espeholt2018impala}.
\item Agents receive the same screen observations that a human playing the game would without access to internal game state, rewards, or any other privileged information \citep[cf.,][]{berner2019dota,vinyals2019grandmaster}.
\item To interact with the environments, agents use the same keyboard-and-mouse controls that humans do \citep[e.g.,][]{baker2022video,humphreys2022data,lifshitz2023steve}, rather than handcrafted action spaces or high-level APIs.
\item We focus on following language instructions \citep[e.g.,][]{abramson2020imitating} rather than simply playing the games to maximize a win-rate or generating plausible behavior \citep[cf.,][]{berner2019dota,vinyals2019grandmaster}.
\item We train and test our agents using open-ended natural language, rather than simplified grammars or command sets \citep[e.g.,][]{abramson2020imitating}.
\end{itemize}

These design choices make the learning problem harder, but their generality makes expanding to new environments easier: agents use the same interface across environments without requiring a custom design of control and observation spaces for each new game. 
Furthermore, since the agent-environment interface is human compatible, it allows agents the potential to achieve anything that a human could, and allows direct imitation learning from human behavior. This general interface from language instructions to embodied behavior can also enable agents to transfer previously learned skills zero-shot to never-before-seen games. Doing research in generic virtual environments allows us to test our agents in a broad and challenging range of situations---where the lessons learned are likely to be more applicable to real-world applications with visually rich perception and control such as robotics---without the risks and costs of real-world testing: if the agent crashes a spaceship in a video game, we can just restart the game.

In the SIMA project thus far, we have created an agent that performs short-horizon tasks based on language instructions produced by a user; though instructions could also be produced by a language model \citep[e.g.,][]{jiang2019language,driess2023palm,wang2023jarvis,hu2023look,ajay2023compositional}. We have a portfolio of over ten 3D environments, consisting of research environments and commercial video games. For research environments we evaluate agents using the ground truth state, but commercial video games are not designed to report on the completion of arbitrary language tasks. We have therefore developed a variety of methods for evaluation in video games, including using optical character recognition (OCR) to detect onscreen text describing task completion, and using human evaluation of recorded videos of agent behavior.
In the rest of this tech report, we describe the high-level approach (illustrated in \Cref{fig:overview}) and our initial progress towards the ultimate goal of SIMA: developing an instructable agent that can accomplish anything a human can do in any simulated 3D environment.

\section{Related work}
\label{sec:related}

SIMA builds on a long history of using games as a platform for AI research. For example, backgammon provided the initial proving ground for early deep reinforcement learning methods \citep{tesauro1995temporal}, and later works have achieved superhuman performance even in complex board games like Go \citep{silver2016mastering,silver2018general}.

\paragraph{Video games}
Over the last ten years, video games have provided an increasingly important setting for research focused on embodied agents that perform visuomotor control  in rich environments. Researchers have used many video game environments, covering a wide spectrum from Atari \citep{bellemare2013arcade} to DoTA \citep{berner2019dota} and StarCraft II \citep{vinyals2019grandmaster}.
In SIMA, however, we restrict our focus to games that resemble 3D physical embodiment most closely, in particular games where the player interacts with a 3D world from a first or over-the-shoulder pseudo-first-person view.
This focus excludes many of the games which have previously been used for research, such as the ones listed above.
There has however been notable interest in first-person embodied video games as a platform for AI research \citep{johnson2016malmo,tessler2017deep,guss2019minerl,pearce2022counter,hafner2023mastering,durante2024interactive,tan2024towards}. These video game AI projects have driven the development of many innovative techniques, e.g., learning from videos by annotating them with estimated player keyboard-and-mouse actions using inverse dynamics models \citep{pearce2022counter,baker2022video}. More recently, games that offer API access to the environment have served as a platform for grounding large language models \citep{wang2023voyager}, and some works have even considered grounding a language model in a game through direct perception and action of a lower-level controller \citep{wang2023jarvis}.
Instead of focusing on a single game or environment, however, SIMA considers a range of diverse games to train agents on a larger variety of content.

\paragraph{Research environments}
Other works have focused on custom, controlled environments designed for research. Many of these environments focus on particular domains of real-world knowledge.
For example, AI2-THOR \citep{kolve2017ai2}, VirtualHome \citep{puig2018virtualhome}, ProcTHOR \citep{deitke2022procthor}, AI Habitat \citep{savva2019habitat,szot2022habitat,puig2023habitat}, ALFRED \citep{shridhar2020alfred}, and Behavior \citep{srivastava2021behavior} simulate embodied agents behaving in naturalistic rendered scenes. CARLA \citep{dosovitskiy2017carla} provides a simulator for autonomous driving. MuJoCo \citep{todorov2012mujoco}, PyBullet \citep{coumans2016}, and Isaac Gym \citep{makoviychuk2021isaac} provide high quality physics simulators for learning low-level control and are used by benchmarks for robotic manipulation such as Meta-World \citep{yu2020meta} and Ravens \citep{zeng2022transporter}. \citet{albrecht2022avalon} propose a unified environment encompassing a variety of skills afforded through ecologically-inspired interactions.
The Playhouse \citep{abramson2020imitating,team2021creating,abramson2022improving} and WorldLab \citep[e.g.,][]{paine2019making} environments are built using Unity \citep[see][]{ward2020using}.
\cite{team2021open} and \cite{team2023human} also use Unity to instantiate a broad distribution of procedurally generated tasks with shared underlying principles.
For the results in this work, we also use Playhouse, WorldLab, and ProcTHOR.
In addition, we introduce a new environment, called the Construction Lab.

\paragraph{Robotics}
Robotics is a key area for research in embodied intelligence. A variety of robotics projects have used simulations for training, to transfer efficiently to real-world robotic deployments \citep{hofer2021sim2real}, though generally within a single, constrained setting. More recent work has focused on environment-generality, including scaling robotic learning datasets across multiple tasks and embodiments  \citep{brohan2022rt1,brohan2023rt2,stone2023open,padalkar2023open}---thereby creating Vision-Language-Action (VLA) models \citep{brohan2023rt2}, similar to the SIMA agent.
The latter challenge of generalizing or quickly adapting to new embodiments has some parallels to acting in a new 3D environment or computer game where the mechanics are different.
Moreover, a variety of recent works have applied pretrained (vision-)language models as a planner for a lower-level instruction-conditional robotic control policy \citep{brohan2023can,driess2023palm,vemprala2023chatgpt,hu2023look}.
Our approach shares a similar philosophy to the many works that attempt to ground language via robotics. SIMA, however, avoids the additional challenges of costly hardware requirements, resource-intensive data collection, and the practical limitations on diversity of real-world evaluation settings.
Instead, SIMA makes progress towards embodied AI by leveraging many simulated environments and commercial video games to obtain the sufficient breadth and richness that we conjecture to be necessary for effectively scaling embodied agents---with the hope that lessons learned (and possibly even the agents themselves) will be applicable to robotic embodiments in the future.

\paragraph{Learning environment models}
Some works attempt to leverage learned models of environments to train agents in these learned simulations \citep[e.g.,][]{ha2018world,hafner2020mastering,hafner2023mastering,yang2023learning}. These methods, however, tend to be difficult to scale to diverse sets of visually complex environments that need to be self-consistent across long periods of time. Nevertheless, learning imperfect models can still be valuable. In SIMA, we build on video models \citep{villegas2022phenaki}, which we fine-tune on game environments. However, we only use the internal state representations of the video models rather than explicit rollouts---in keeping with other approaches that use generative modeling as an objective function for learning state representations \citep[e.g.,][]{gregor2019shaping,zolna2024gats}.

\paragraph{Grounding language}
Another stream of work---overlapping with those above---has focused on grounding language in simulated 3D environments, through agents that are trained in controlled settings with semi-natural synthetic language \citep{hermann2017grounded,hill2019environmental}, or by imitating human interactions in a virtual house to learn a broader ability to follow natural language instructions \citep{abramson2020imitating,team2021creating,abramson2022improving,abramson2022evaluating}. 
Moreover, a range of recent works develop agents that connect language to embodied action, generally as part of a hierarchy controlled by a language model \citep{jiang2019language,driess2023palm,wang2023jarvis,hu2023look,ajay2023compositional}. We likewise draw inspiration from the idea that language is an ideal interface for directing an agent, but extend our scope beyond the limited affordances of a single controlled environment.
In that sense, SIMA overlaps more with several recent works \citep{reed2022generalist,huang2023embodied,durante2024interactive} that also explore training a single model to perform a broad range of tasks involving actions, vision, and language.
However, SIMA is distinct in our focus on simultaneously (1) taking a language-first perspective, with all training experiences being language-driven; (2) adopting a unified, human-like interface across environments with language and vision to keyboard-and-mouse control; and (3) exploring a broad range of visually rich, diverse, and human-compatible environments that afford a wide range of complex skills.

\paragraph{Language supports grounded learning, and grounded learning supports language} 
A key motivation of SIMA is the idea that learning language and learning about environments are mutually reinforcing. A variety of studies have found that even when language is not \emph{necessary} for solving a task, learning language can help agents to learn generalizable representations and abstractions, or to learn more efficiently. Language abstractions can accelerate grounded learning, for example accelerating novelty-based exploration in reinforcement learning by providing better state abstractions \citep{tam2022semantic,mu2022improving}, or composing known goals into new ones \citep{colas2020language,nottingham2023embodied}. Moreover, learning to predict natural-language explanations \citep{lampinen2022tell}, descriptions \citep{kumar2022using}, or plans \citep{hu2023thought} can help agents to learn more efficiently, and to generalize better out of distribution. Language may be a powerful tool for shaping agent capabilities \citep{colas2022language}.

Conversely, richly grounded learning can also support language learning. Since human language use is deeply integrated with our understanding of grounded situations \citep{mcclelland2020placing}, understanding the subtleties of human language will likely benefit from this grounding. Beyond this theoretical argument, empirical evidence shows that grounding can support even fundamental kinds of generalization---\citet{hill2019environmental} show that agents grounded in richer, more-embodied environments exhibit more systematic compositional generalization. These findings motivate the possibility that learning both language and its grounding will not only improve grounded actions, but improve a system's knowledge of language itself.

\section{Approach} 
\label{sec:approach}

Many overlapping areas of previous and concurrent work share some of our philosophy, motivations, and approaches. What distinguishes the SIMA project is our focus on language-conditional behavior across a diverse range of visually and mechanically complex simulated environments that afford a rich set of skills. In this section, we provide a high-level overview of our approach: our environments, data, agents, and evaluations.

\subsection{Environments}
\label{sec:environments}

\begin{figure*}
    \centering
    \includegraphics[width=\linewidth]{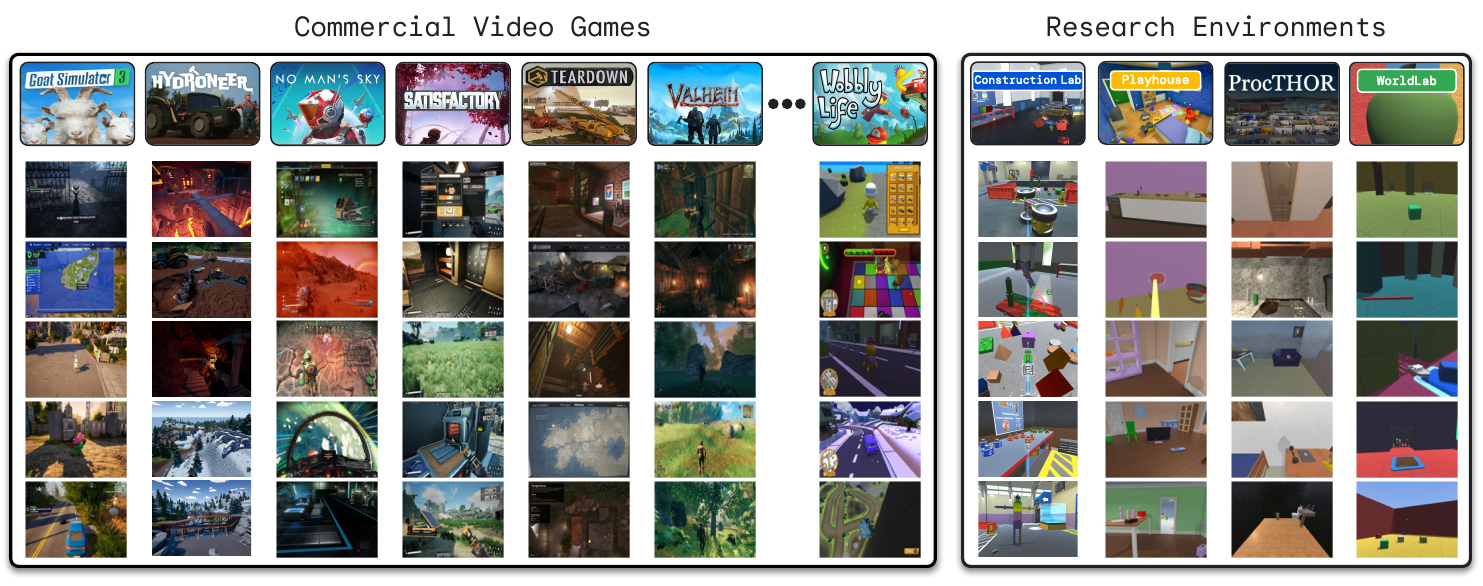}
    \caption{\textbf{Environments.} We use over ten 3D environments in SIMA, consisting of commercial video games and research environments. The diversity of these environments is seen in their wide range of visual observations and environmental affordances. Yet, because these are all 3D environments, basic aspects of 3D embodied interaction, such as navigation, are shared. Commercial video games offer a higher degree of rich interactions and visual fidelity, while research environments serve as a useful testbed for probing agent capabilities.}
    \label{fig:environments}
\end{figure*}

SIMA aims to ground language across many rich 3D environments (\Cref{fig:environments}). Thus, we selected 3D embodied environments that offer a broad range of open-ended interactions---such environments afford the possibility of rich and deep language interactions. We focus on environments that are either in a) first-person or b) third-person with the camera over the player's shoulder. To achieve diversity and depth of experience, we use a variety of commercial video games, as well as several environments created specifically for agent research. Each type of environment offers distinct advantages, ranging from open-ended diverse experiences to targeted assessments of agent skills. We have deliberately sought to build a portfolio of games that covers a wide range of settings---from mundane tasks in semi-realistic environments, to acting as a mischevious goat in a world with exaggerated physics, to exploring mythological worlds or science-fiction universes. Below, we briefly describe the environments we have used in SIMA thus far by category and in alphabetical order.

\subsubsection{Commercial video games}

Commercial video games offer exciting, open-ended worlds full of visual richness and the potential for complex interactions. In SIMA, we have partnered with games developers whose games we used for training agents, and we are continuing to develop relationships with new developers---for our full list of current partners, please see our Acknowledgements section. We focus on a variety of open-world or sandbox games that contain diverse skills, while avoiding games containing harmful content such as extreme violence or biases. We have also sought a broad diversity of worlds and stories, but with a focus on games that exhibit a depth of interesting mechanics. Accordingly, games from our portfolio offer a wide range of distinct challenges in perception and action, from flying a spaceship to mining minerals or crafting armor, as well as more common core features, such as navigation or gathering resources. Games also often include interactions that extend beyond the skillset of typical embodied research environments, such as menu use and interfaces more similar to those faced in computer control benchmarks \citep[e.g.,][]{humphreys2022data,koh2024visualwebarena}. For the results in this report, we focus on single-player interactions within these games.

We run instances of each game in a secure Google Cloud environment, using hardware accelerated rendering to a virtual display. This display is streamed to a browser for human gameplay, or to a remote agent client process during evaluation. To instantiate repeatable evaluation or data collection scenarios within each game, we build datasets of save-game files from expert play, and use scripted processes to automate the process of installing game-files, booting the game, navigating its main menu, and loading a specific save-game.

We now provide a brief description of the games we used.

\textbf{Goat Simulator 3:} A third-person game where the player is a goat in a world with exaggerated physics. The player can complete quests, most of which involve wreaking havoc. The goat is able to lick, headbutt, climb, drive, equip a wide range of visual and functional items, and perform various other actions. Throughout the course of the game, the goat unlocks new abilities, such as the ability to fly.

\textbf{Hydroneer:} A first-person mining and base building sandbox where the player is tasked with digging for gold and other resources to turn a profit and enhance their mining operation. To do this, they must build and upgrade their set-ups and increase the complexity and levels of automation until they have a fully automated mining system. Players can also complete quests from non-player characters to craft bespoke objects and gain extra money. Hydroneer requires careful planning and managing of resources.

\textbf{No Man's Sky:} A first- or third-person survival game where the player seeks to explore a galaxy full of procedurally-generated planets. This involves flying between planets to gather resources, trade, build bases, and craft items that are needed to upgrade their equipment and spaceship while surviving a hazardous environment. No Man's Sky includes a large amount of visual diversity---which poses important challenges for agent perception---and rich interactions and skills.

\textbf{Satisfactory:} A first-person, open-world exploration and factory building game, in which players attempt to build a space elevator on an alien planet. This requires building increasingly complex production chains to extract natural resources and convert them into industrial goods, tools, and structures---whilst navigating increasingly hostile areas of a large open environment.

\textbf{Teardown:} A first-person, sandbox–puzzle game in a fully destructible voxel world where players are tasked with completing heists to gain money, acquiring better tools, and undertaking even more high-risk heists. Each heist is a unique scenario in one of a variety of locations where players must assess the situation, plan the execution of their mission, avoid triggering alarms, and escape before a timer expires. Teardown involves planning and using the environment to one's advantage to complete the tasks with precision and speed.

\textbf{Valheim:} A third-person survival and sandbox game in a world inspired by Norse mythology. Players must explore various biomes, gather resources, hunt animals, build shelter, craft equipment, sail the oceans and defeat mythological monsters to advance in the game---while surviving challenges like hunger and cold.

\textbf{Wobbly Life:} A third-person, open-world sandbox game where the player can explore the world, unlock secrets, and complete various jobs to earn money and buy items, leading up to buying their own house. They must complete these jobs whilst contending with the rag-doll physics of their characters and competing against the clock. The jobs require timing, planning, and precision to be completed. The world is extensive and varied, with a diverse range of interactive objects.

\subsubsection{Research environments}

In contrast to commercial video games, AI research environments are typically more controllable, offering the ability to instill and carefully assess particular skills, and more rapid and reliable evaluations of task completion. Unlike many of the games in our portfolio, several of these research environments also tend to feature more real-world analogous---if still simplified---physical interactions.

We have drawn on several prior research environments and developed a new environment---the Construction Lab---that incorporates important challenges which were not otherwise well-captured by our other environments. 

\textbf{Construction Lab:} A new research environment where agents need to build novel items and sculptures from interconnecting building blocks, including ramps to climb, bridges to cross, and dynamic contraptions. Construction Lab focuses on cognitive capabilities such as object manipulation and an intuitive understanding of the physical world.

\textbf{Playhouse:} An environment used in various prior works \citep{abramson2020imitating,team2021creating,abramson2022improving}, consisting of a procedurally-generated house environment with various objects. We have augmented this environment with improved graphics and richer interactions, including skills like cooking or painting.

\textbf{ProcTHOR:} An environment consisting of procedurally-generated rooms with realistic contents, such as offices and libraries, introduced by \cite{deitke2022procthor}. Although benchmark task sets exist in this environment, prior works have not used keyboard and mouse actions for agents; thus we focus on this environment primarily for data collection rather than evaluation.

\textbf{WorldLab:} An environment used in prior work \citep[e.g.,][]{paine2019making}, further specialized for testing embodied agents by using a limited set of intuitive mechanics, such as sensors and doors, and relying primarily on the use of simulated physics on a range of objects.

\subsection{Data}
\label{sec:data}

\begin{figure}
    \centering
    \includegraphics[width=0.8\linewidth]{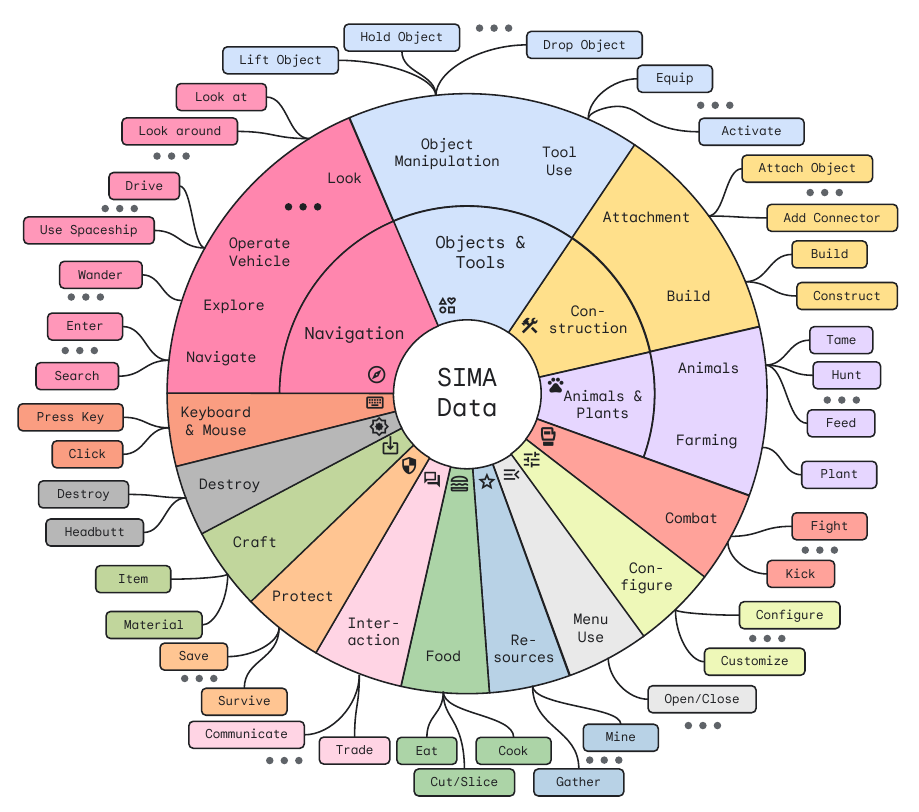}
    \caption{\textbf{Instructions Across SIMA Data.} The SIMA dataset includes a broad range of text instructions that can be roughly clustered into a hierarchy. Due to the common 3D embodied nature of the environments that we consider, many generic tasks, such as navigation and object manipulation, are present in multiple environments. Categories were derived from a data-driven hierarchical clustering analysis of the human-generated text instructions within a fixed, pretrained word embedding space. Note that the area of each cluster in the wheel in \Cref{fig:skills} does not correspond to the exact number of instructions from that cluster in the dataset.}
    \label{fig:skills}
\end{figure}

Our approach relies on training agents at scale via behavioral cloning, i.e., supervised learning of the mapping from observations to actions on data generated by humans. Thus, a major focus of our effort is on collecting and incorporating gameplay data from human experts. This includes videos, language instructions and dialogue, recorded actions, and various annotations such as descriptions or marks of success or failure. These data constitute a rich, multi-modal dataset of embodied interaction within over 10 simulated environments, with more to come.\footnote{Note: Due to a limited amount of collected data and/or evaluations, we present agent evaluation results (\Cref{sec:results}) on a subset of 7 of these environments.} Our data can be used to augment and leverage existing training data \citep[e.g.,][]{abramson2020imitating}, or to fine-tune pretrained models to endow them with more situated understanding. These datasets cover a broad range of instructed tasks: \Cref{fig:skills} shows instruction clusters derived from hierarchically clustering the text instructions present in the data within a fixed, pretrained word embedding space. 

Yet, collecting data at scale is not sufficient for training successful agents. Data quality processes are critical to ensuring an accurate and unconfounded mapping between language and behavior. This presents various technical challenges. We take care to engineer our data collections, including preprocessing and filtering the raw data, to highlight important skills and effectively train our agents.

\paragraph{Data collections} We collect data using a variety of methods, including allowing single players to freely play, and then annotating these trajectories with instructions post-hoc. We also perform two-player setter-solver collections \citep{abramson2020imitating,team2021creating}, in which one player instructs another what to do in selected scenarios while sharing a single player view in order to match the single-player collections. All our data collections were performed with participants contracting with Google. The full details of our data collection protocols, including compensation rates, were reviewed and approved by an independent Human Behavioral Research Committee for ethics and privacy. All participants provided informed consent prior to completing tasks and were reimbursed for their time.

\paragraph{Preprocessing, filtering, and weighting} Before training, we perform a variety of offline preprocessing steps, including resizing data for agent input, filtering out low-quality data using a variety of heuristics, and remixing and weighting data across environments and collections to prioritize the most effective learning experiences.

\subsection{Agent}
\label{sec:agents}

\begin{figure*}
    \centering
    \includegraphics[width=\linewidth]{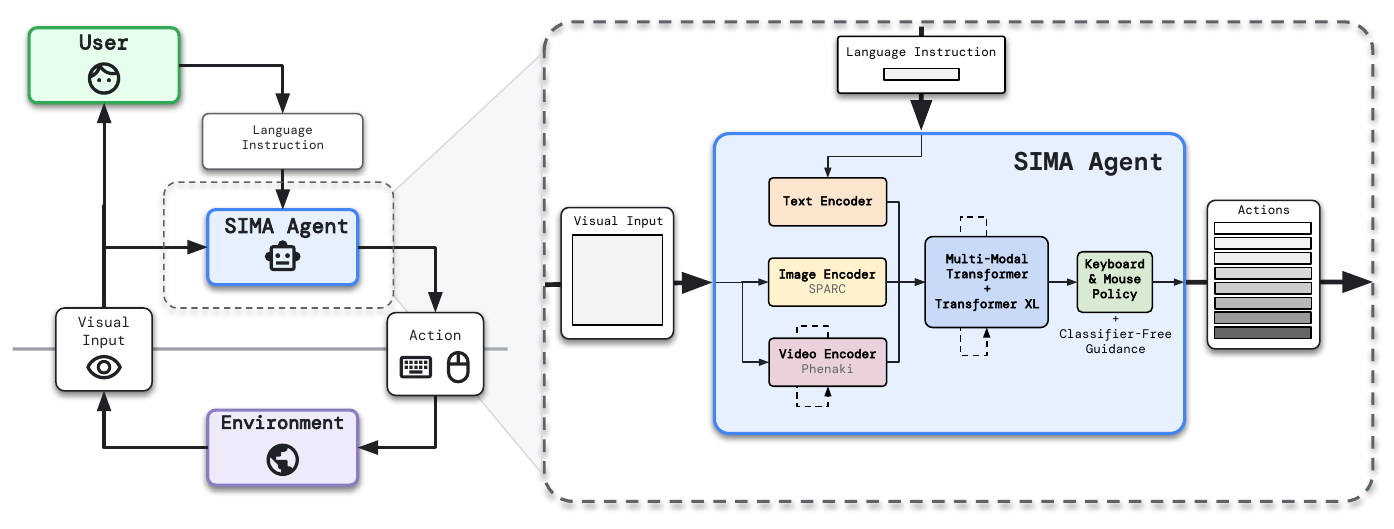}
    \caption{\textbf{Setup \& SIMA Agent Architecture.} The SIMA agent receives language instructions from a user and image observations from the environment, and maps them to keyboard-and-mouse actions.}
    \label{fig:agent}
\end{figure*}

The SIMA agent maps visual observations and language instructions to keyboard-and-mouse actions (\Cref{fig:agent}). Given the complexity of this undertaking---such as the high dimensionality of the input and output spaces, and the breadth of possible instructions over long timescales---we predominantly focus on training the agent to perform instructions that can be completed in less than approximately 10 seconds. Breaking tasks into simpler sub-tasks enables their reuse across different settings and entirely different environments, given an appropriate sequence of instructions from the user.

Our agent architecture builds on prior related work \citep{abramson2020imitating,abramson2022improving}, but with various changes and adaptations to our more general goals. First, our agent incorporates not only trained-from-scratch components, but also several pretrained models---including a model trained on fine-grained image-text alignment, SPARC \citep{bica2024improving}, and a video prediction model, Phenaki \citep{villegas2022phenaki}---which we further fine-tune on our data through behavioral cloning and video prediction, respectively. In preliminary experiments, we found that these models offer complementary benefits. Combining these pre-trained models with fine-tuning and from-scratch training allows the agent to utilize internet-scale pretraining while still specializing to particular aspects of the environments and the control tasks that it encounters.

More specifically, our agent (\Cref{fig:agent}) utilizes trained-from-scratch transformers that cross-attend to the different pretrained vision components, the encoded language instruction, and a Transformer-XL \citep{dai2019transformer} that attends to past memory states to construct a state representation. The resulting state representation is provided as input to a policy network that produces keyboard-and-mouse actions for sequences of 8 actions. We train this agent with behavioral cloning, as well as an auxiliary objective of predicting goal completion.

We use Classifier-Free Guidance (CFG; \citealp{ho2022classifier,lifshitz2023steve}) to improve the language-conditionality of a trained agent when running it in an environment. CFG was originally proposed for strengthening text-conditioning in diffusion models \citep{ho2022classifier}, but has also proven useful for similar purposes with language models \citep{sanchez2023stay} and language-conditioned agents \citep{lifshitz2023steve}. That is, we compute the policy, \(\pi\), with and without language conditioning, and shift the policy logits in the direction of the difference between the two:

\[
\pi_{CFG} = \pi\left(\text{image}, \text{language}\right) + \lambda \left(\pi\left(\text{image}, \text{language}\right) - \pi\left(\text{image}, \cdot \right)\right).
\]

\subsection{Evaluation methods}
\label{sec:evaluation}

Our focus on generality in SIMA introduces challenges for evaluation. While research environments may provide automated methods for assessing whether language-following tasks have been successfully completed, such success criteria may not be generally available. That is, language instructions may not correspond to goal states recorded by an environment (e.g. a user might instruct \textit{``make a pile of rocks to mark this spot''} or \textit{``see if you can jump over this chasm''}).

Evaluating agents in commercial video games poses substantial additional challenges. Video game evaluations cannot rely on access to privileged information about the state of an environment. Additionally, it is difficult to reinstate agents in precisely the same state in environments that are not designed as reproducible benchmarks, and loading each task in commercial video games is considerably slower and more costly than those in research environments. Achieving fast, stable, and reliable evaluations comparable across environments is thus challenging. We therefore use a range of distinct evaluation types that provide different trade-offs in efficiency, cost, accuracy, and coverage.

Moreover, ensuring that our evaluations truly assess language conditionality, rather than environmental affordances, requires care. For instance, if a task contains a knife, a cutting board, and a carrot, the agent may ascertain the goal (\textit{``cut the carrot on the cutting board''}) without relying on the language instruction. Thus, task settings need to afford a diversity of actions, ideally testing multiple instructions from a single initial state, to properly evaluate whether the agent's actions are driven by language.

\paragraph{Action log-probabilities}
One simple approach is to evaluate agents based on their action predictions on held-out evaluation data. However, consistent with prior findings \citep{abramson2022evaluating,baker2022video}, we observed that agent action log-probabilities on evaluation data show at most a weak correlation with agent performance beyond the most basic skills. Thus, online evaluations, in which the agent interacts with the environment, are needed to understand agent performance in detail.

\paragraph{Static visual input}
Similar to predicting actions on held-out data, we can provide the agent with a static visual input and a language instruction to perform a particular valid action (e.g., \textit{``jump''}) to assess simple responses directly mapping to particular keyboard and/or mouse actions. We have used evaluations of this form for our commercial video game environments, as they have the advantage of not requiring actually loading a game. While these evaluations can be a useful early signal, they do not reliably predict success on prolonged tasks.

\paragraph{Ground-truth}
Our internally-developed research environments (Construction Lab, Playhouse, and WorldLab) are capable of providing ground-truth assessments of whether language-following tasks have been successfully completed. These tasks can depend on the state of the agent (\textit{``move forward''}) and the surrounding environment (\textit{``lift the green cube''}), as well as more complex interactions (\textit{``attach a connector point to the top of the large block''} or \textit{``use the knife to chop the carrots''}). Such tasks enable robust testing of a range of particular skills, with a highly reliable signal of task success. Moreover, we design the task settings and evaluation to be strong tests of precision; for example, many tasks include distractor objects, for which the episode is marked as an immediate failure if the agent interacts with the distractors rather than the instruction target---even if the agent might have completed the actual task later. We also include other types of assessments, such as instructing the agent to complete one goal, and then interrupting with another goal to evaluate whether it switches appropriately---this ensures that agents are sufficiently responsive to changes in commands. A subset of our research environment tasks are used to provide a fast evaluation signal of agent progress during training.

\paragraph{Optical character recognition (OCR)}
Many of our commercial video game environments provide on-screen text signalling the completion of tasks or quests, or even the results of lower-level actions like collecting resources or entering certain areas of a game.
By detecting on-screen text using OCR in pre-defined evaluation scenarios, sometimes in combination with detecting specific keyboard-and-mouse actions, we can cheaply assess whether the agent has successfully performed particular tasks.
This form of automated evaluation also avoids the subjectivity of human evaluations.
We make use of OCR evaluation in particular for two games, No Man's Sky and Valheim, which both feature a significant amount of on-screen text.
In No Man's Sky, for example, we have developed evaluation tasks such as \textit{``mine carbon/salt/ferrite''}, \textit{``use the analysis visor''}, or \textit{``open the exosuit menu''}.
Similarly, in Valheim we have tasks such as \textit{``collect wood/stone/raspberries''}, \textit{``use the workbench''}, or \textit{``cook food''}.
In general, however, OCR evaluations are restricted to tasks that signal completion with game-specific text rather than arbitrary tasks that can be specified with language instructions and which we would expect a general agent to be able to solve.
Other video games also have significantly less on-screen text, which makes the range of behaviors that can be evaluated in these games with OCR very narrow.

\paragraph{Human evaluation}
In the many cases where we cannot automatically derive a signal of task success, we turn to humans to provide this assessment. While this is our most general evaluation method, it is also the slowest and most expensive. We use human judges who are game experts, i.e., they have played these specific games for at least 16 hours, and often over the course of several weeks. We ask them to review recorded agent videos, collecting multiple ratings of the same video from different judges (typically 5) to ensure reliable assessments. We also encourage strict evaluations: we instruct judges to mark an episode as a failure in cases where the agent performs irrelevant actions first, even if the agent successfully completes the instructed task afterward.

We curated our human-evaluation tasks by identifying a list of frequently-occurring verbs in English, and combined it with a list of verbs that naturally emerged from gameplay and interactive testing of our agents. We use this verb list as a foundation for our evaluations across all video game environments. We assign each task (save state and instruction pair) to a single, most-representative skill category (e.g. ``craft items''), even though most tasks require a wide range of implicit skills to succeed (e.g. crafting often requires menu use). The resulting evaluation set provides a long term challenge for agent research that spans a wide range of difficulties---from simple game agnostic tasks such as \textit{``turn left''}, to ones testing specialized game knowledge \textit{``compare the crafting cost of antimatter and antimatter housing''}, to ones utilising broader semantic knowledge such as \textit{``take the pitchfork from the person shoveling hay''}. Grounding our evaluation framework in the distribution of natural language allows us to test our agents in both common and adversarial scenarios, and thereby to measure our progress towards our long-term goal of developing an instructable agent that can accomplish anything a human can do in any simulated 3D environment.

In the results below (\Cref{sec:results}), we primarily report evaluation scores based on ground-truth evaluations for research environments and combined OCR and human evaluations for commercial video game environments. Across the 7 environments for which we have evaluations, we have a total of 1,485 unique tasks, spanning a range of 9 skill categories, from movement (\textit{``go ahead'', ``look up'', ``jump''}) to navigation (\textit{``go to the HUB terminal'', ``go to your ship''}), resource gathering (\textit{``collect carbon'', ``get raspberries''}), object management (\textit{``use the analysis visor'', ``cut the potato''}), and more.
(For reference, MineDojo \citep{fan2022minedojo}, a related work investigating language-conditional agents in MineCraft, used 1,581 unique tasks spanning 4 skill categories: survival, harvest, tech-free, and combat).  Given the diversity and coverage of our current evaluations, they provide a reasonable assessment of the fundamental language-conditional skills that we expect from our agent. Yet, there remains ongoing work in developing more scalable, general, and reliable evaluations, particularly as we move toward more complex and open-ended tasks.

\subsubsection{Latency mitigations}

Our agent is evaluated in several environments that run in real-time, asynchronously to the agent. This can pose challenges for the timely execution of agent-generated actions. Latencies or delays \citep{bratko1995behavioural} are introduced by the computation of actions and the transmission of observations and actions over the network. We account for this latency during behavioral cloning by predicting actions that are offset in time relative to the visual input to the agent, and mirror this offset during evaluation by appropriate buffering of observations and actions during neural-network inference. We additionally minimize latencies with appropriate scheduling of action computation on TPU accelerators, on-device caching of neural-network state across timesteps, and by careful choices of batch size and other implementation details.

\subsection{Responsibility}
We follow a structured approach to responsible model development, to identify, measure, and manage foreseeable ethics and safety challenges. These are informed by academic literature reviews, engaging with internal ethics teams, and developing comprehensive ethical assessments that document key risks with mitigation strategies. We ensure that our research projects uphold Google’s AI Principles.\footnote{\url{https://ai.google/responsibility/principles/}}  SIMA was carefully assessed and reviewed to ensure that its societal benefits outweigh the risks, and that appropriate risk mitigations are incorporated.

\paragraph{Benefits}
SIMA is a cutting-edge research initiative which focuses on how to develop instructable agents in simulated environments. This research presents interesting opportunities for the future of humans and AI collaborating together; unlike LLMs, SIMA is able to both understand natural language instructions and dynamic, interactive 3D environments. This presents a new paradigm for working with AI agents, and the potential for exciting new immersive 3D experiences with AI. Finally, simulated environments present a safer alternative for research compared to other AI deployments.

\paragraph{Risks}
As well as these benefits, we have reflected on potential risks associated with training on video game data. 
These include risks associated with training an agent on games that include violent, explicit or otherwise harmful behaviors. We have also reflected on the implications on representational harms, as the agent may learn from stereotyped depictions or actions in game settings. 
Besides these risks, there are also down stream risks associated with the future hypothetical deployments of SIMA, through either intentional misuse or benign action. 

\begin{figure*}
    \centering
    \includegraphics[width=\textwidth]{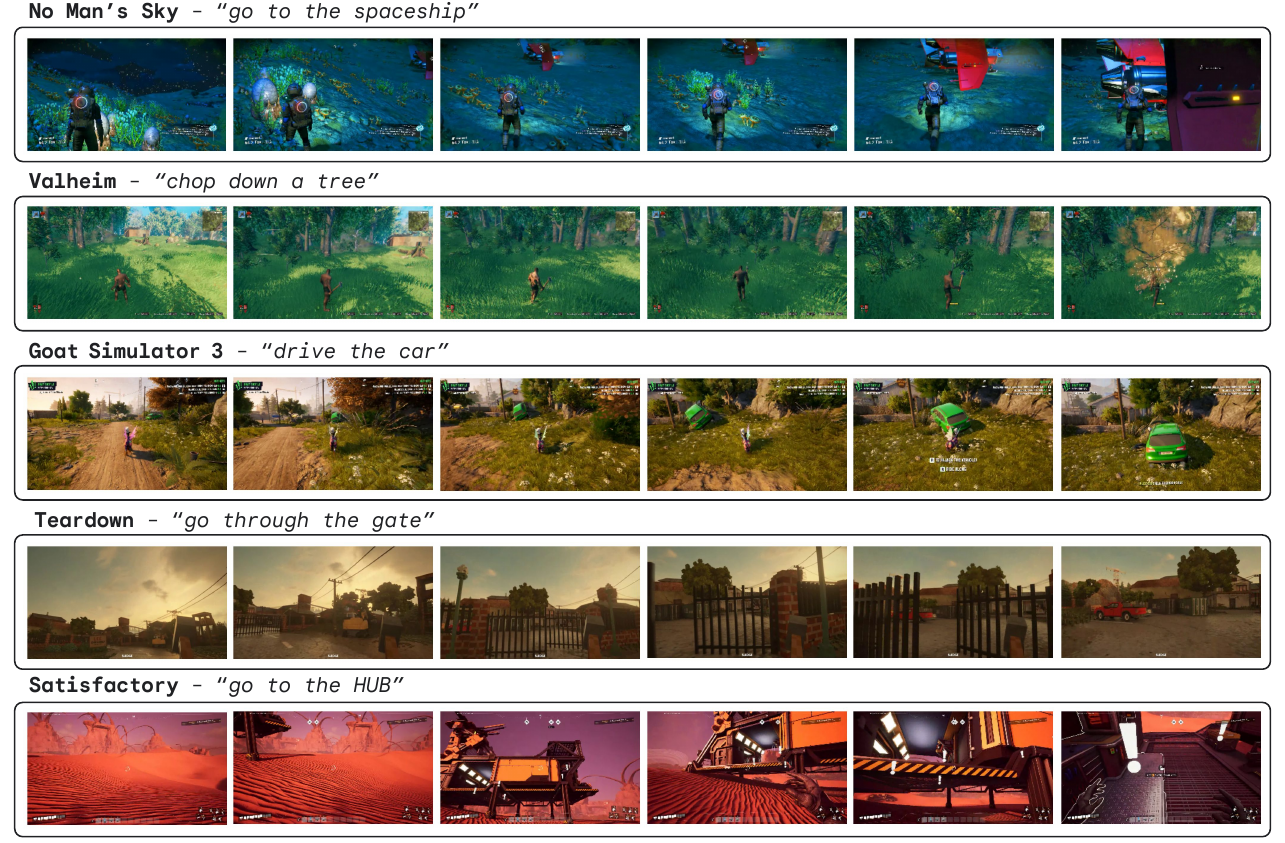}
    \caption{\textbf{Agent Trajectories.} The SIMA agent is capable of performing a range of language-instructed tasks across diverse 3D virtual environments. Here, we provide several representative, visually salient examples of the agent's capabilities that demonstrate basic navigation and tool use skills.}
    \label{fig:agent trajectories}
\end{figure*}

\paragraph{Mitigations}
We have worked to ameliorate these risks through a holistic approach, including:
\begin{itemize}[topsep=0pt,noitemsep]
\item Careful curation of content. We avoided a number of games that have scientifically interesting, but violent environments. We also outlined behavioral ``red-lines'' with our ethics and safety teams; games with content that violates these red-lines are not used.
\item Continuous evaluations of SIMA’s safety performance.
\item Ensuring SIMA’s deployments and agreements are transparent, and for now remain in a controlled, closed environment. 
\end{itemize}
Ultimately, given the careful training data selection and constrained deployment environment of SIMA, we are confident we can maximize the benefits while minimising the ethical risks.

\section{Initial results}
\label{sec:results}

In this section, we report initial evaluation results of the SIMA agent. After presenting several qualitative examples of the SIMA agent's capabilities, we start by considering the quantitative performance of the SIMA agent, broken down by environment and skill category. We then compare these results with several baselines and ablations, allowing us to assess the generalization capabilities of the agent and the efficacy of our design choices. Finally, we investigate a subset of evaluation tasks to estimate human-level performance as an additional comparison.

\paragraph{Qualitative examples}
To provide a sense of the agent's general capabilities, \Cref{fig:agent trajectories} displays several representative examples of the agent in our commercial video game environments. Despite the visual diversity of the environments, the agent is capable of performing these tasks, demonstrating basic navigation and tool use skills. Even when the instructed target is not in view (\textit{``go to the spaceship''} and \textit{``go to the HUB''}), the agent is able to find the target. For further qualitative examples, please refer to the accompanying website.\footnote{ \scriptsize \url{https://deepmind.google/discover/blog/sima-generalist-ai-agent-for-3d-virtual-environments/}}

\subsection{Performance across environments and skills}

\begin{figure*}
    \centering
    \includegraphics[width=0.97\textwidth]{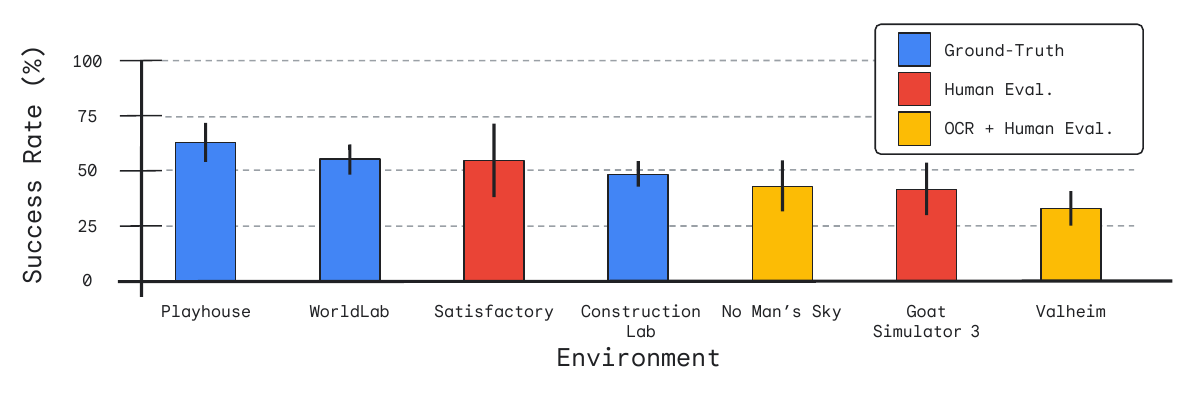}
    \caption{\textbf{Average Success Rate of the SIMA Agent by Environment.} Agents achieve notable success, but are far from perfect; their success rates vary by environment. Colors indicate the evaluation method(s) used to assess performance for that environment. (Note that humans would also find some of these tasks challenging, and thus human-level performance would not be 100\%, see \Cref{sec:results:humans}.)}
    \label{fig:performance_by_domain_absolute}
\end{figure*}

In \Cref{fig:performance_by_domain_absolute}, we report the average performance of the SIMA agent across the seven environments for which we have quantitative evaluations. Averages are calculated across multiple episodes per task (in research environments, one episode per task in video games), multiple tasks per environment, and across three training runs with different random seeds. Error bars denote the 95\% confidence intervals (CIs) across the tasks within that environment and the three training runs with different random seeds.
We note that developing informative evaluation tasks is in itself an ongoing effort, and the quantitative results in this work reflect only the range of particular behaviors that are evaluated at this point in time. 

Overall, the results show that the SIMA agent is able to complete a range of tasks across many environments, but there remains substantial room for improvement. Performance is better for Playhouse and WorldLab, which are comparatively simpler research environments. For the more complex commercial video game environments, we see that performance is, understandably, somewhat lower. Notably, performance on Construction Lab is lower as well, highlighting the relative difficulty of this research environment and its evaluation tasks. This enables the SIMA platform to serve as a useful testbed for further development of agents that can connect language to perception and action.

\begin{figure*}
    \centering
    \includegraphics[width=0.95\textwidth]{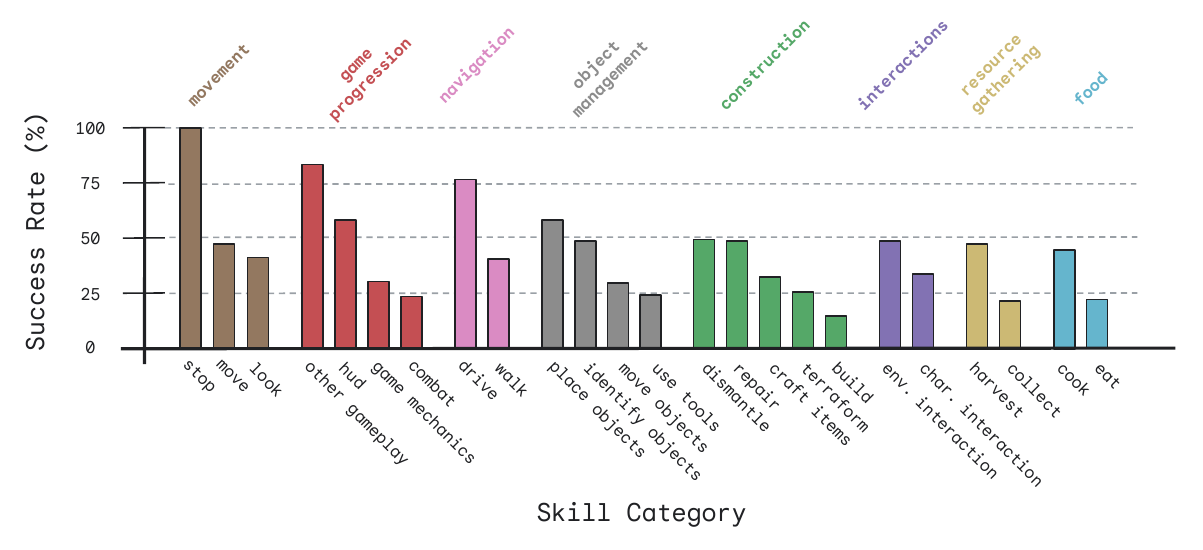}
    \caption{\textbf{Average Success Rate of the SIMA Agent by Skill Category}. Agents exhibit varying degrees of performance across the diverse skills that we evaluate, performing some skills reliably and others with more limited success. Skill categories are grouped into clusters (color), which are derived from our evaluation tasks.}
    \label{fig:performance by skill}
\end{figure*}

In order to better understand the performance of the SIMA agent across an increasing variety of simulated environments, we developed an evaluation framework grounded in natural language for adding and clustering evaluation tasks, as detailed in our evaluation methods. As these skill clusters are derived from our evaluation tasks rather than the training data, they are similar to, yet distinct from, those in \Cref{fig:skills}. As shown in \Cref{fig:performance by skill}, performance varies across different skill categories, including within skill clusters such as ``movement'' or ``game progression''. Note that even seemingly simple skill clusters can involve nontrivial game interactions, e.g., some of the ``look'' tasks involve skills like steering a spaceship (\textit{``look at a planet''}) or orienting based on the surrounding terrain (\textit{``look downhill''}). While there are many subtleties depending on these additional interactions and the mechanics of the environment in which the skill is used, in general, skills that require more precise actions or spatial understanding (``combat'', ``use tools'', ``build'') tend to be more challenging.

\subsection{Evaluating environment generalization \& ablations}

\begin{figure*}
    \centering
    \includegraphics[width=0.85\textwidth]{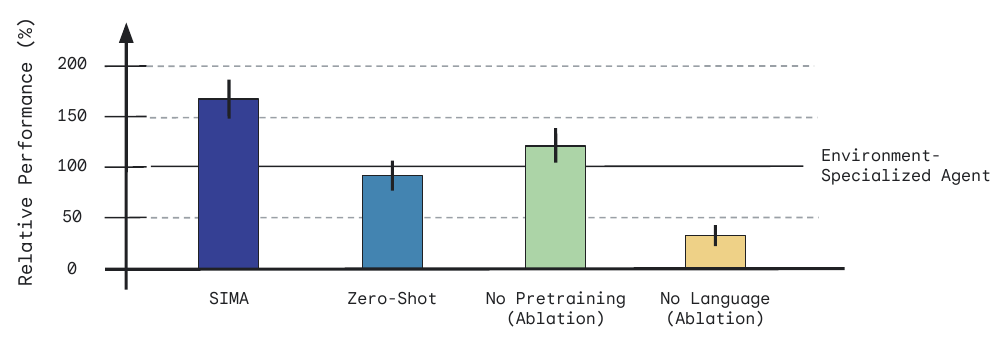}
    \caption{\textbf{Aggregate Relative Performance.} Bars indicate the performance of the SIMA agent as well as the baselines and ablations relative to the performance of the environment-specialized agents, aggregated equally across environments. The SIMA agent outperforms ablations that do not incorporate internet pretraining and substantially outperforms an ablation without language. The solid line shows environment-specialized relative performance, which by normalization is 100\%.}
    \label{fig:performance_summary}
\end{figure*}

\begin{figure*}
    \centering
    \includegraphics[width=\textwidth]{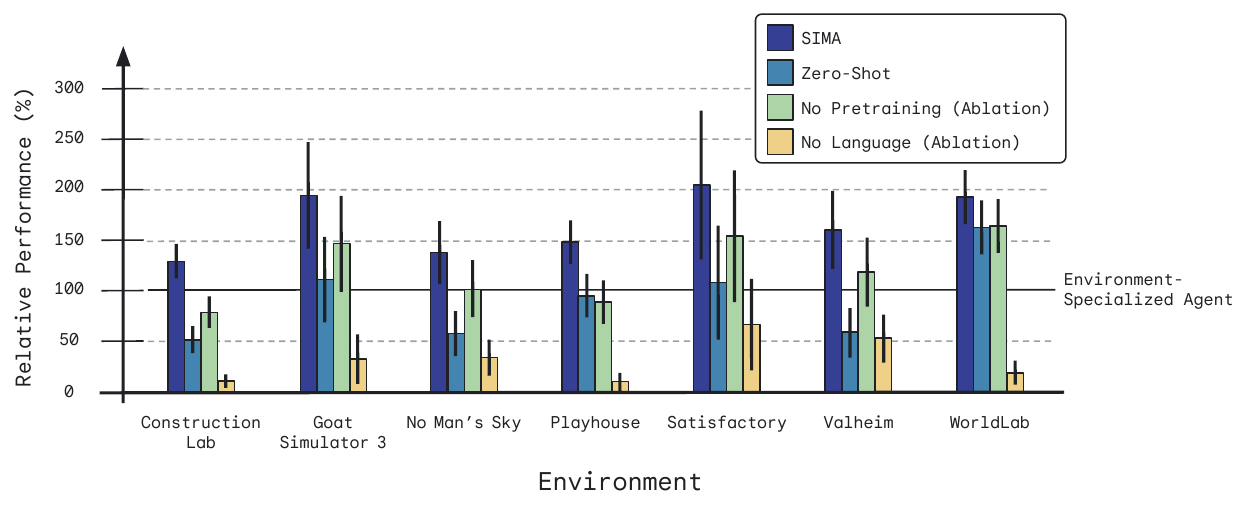}
    \caption{\textbf{Per-Environment Relative Performance.} Bars indicate the performance of the SIMA agent as well as the baselines and ablations relative to the performance of the environment-specialized agents. While performance varies across the environments, the general pattern of results is largely preserved. Even when trained while holding out an environment and evaluated zero-shot on the unseen environment, our agent can achieve non-trivial performance---almost always outperforming the no-language ablation, and in some cases even matching or exceeding environment-specialized agent performance. The solid line shows the relative performance of an environment-specialized agent, which by normalization is 100\%.}
    \label{fig:performance_by_domain}
\end{figure*}

We compare our main SIMA agent to various baselines and ablations, both in aggregate (\Cref{fig:performance_summary}) and broken down across our environments (\Cref{fig:performance_by_domain}). The agents we report across all environments include:
\begin{itemize}
    \item \textbf{SIMA:} Our main SIMA agent, which is trained across all environments except for Hydroneer and Wobbly Life, which we use for qualitative zero-shot evaluation.
    \item \textbf{Zero-shot:} Separate SIMA agents trained like the main agent, but only on \(N-1\) of our environments, and evaluated zero-shot on the held-out environment---that is, without any BC training on it. These agents assess the transfer ability of our agent in a controlled setting. (Note that these agents use the same pretrained encoders as the main SIMA agent, which were finetuned on data from a subset of our environments; thus, in some cases the pretrained encoders will have been tuned with visual inputs from the held-out environment, even though the agent has not been trained to act in that environment. However, the encoders were not fine-tuned on data from Goat Simulator 3, thus the transfer results in that case are unconfounded.)
    \item \textbf{No pretraining ablation:} An agent where we removed the pretrained encoders in the SIMA agent. We replaced these models with a ResNet vision model that is trained from scratch \citep[as in][]{abramson2022improving}, as in preliminary experiments we found training the SPARC/Phenaki encoders through agent training resulted in poor performance. Comparing to this agent tests the benefits of pretrained models for agent performance. 
    \item \textbf{No language ablation:} An agent that lacks language inputs, during training as well as evaluation. Comparing to this agent shows the degree to which our agent's performance can be explained by simple language-agnostic behavioral priors.
    \item \textbf{Environment-specialized:} We additionally train an expert agent on each environment, which is trained only on data corresponding to that environment, but still includes the more broadly pretrained encoders. We normalize the performance of all other agents by the expert agent on each environment, as a measure of what is possible using our methods and the data we have for that environment.
\end{itemize}

Note that due to the number of comparison agents, we only ran a single seed for each, rather than the three seeds used for the main SIMA agent. Each agent is evaluated after 1.2 million training steps.\footnote{With one exception: as we had a relatively small quantity of data for Goat Simulator 3, we attempted to prevent the environment-specialized baseline from overfitting by evaluating it every 200,000 training steps, then selecting the best performing number of steps, which was 400,000 steps, as our environment-specialized baseline. Although this is a biased selection process, because we are using the environment-specialized agent as a baseline, it will only lead to \emph{underestimating} the advantage of SIMA.}
The bars in \Cref{fig:performance_summary} and \Cref{fig:performance_by_domain} represent average performance (normalized relative to the environment-specialist); the errorbars are parametric 95\%-CIs across tasks and seeds (where multiple seeds are available).

\Cref{fig:performance_summary} shows a summary of our results, while \Cref{fig:performance_by_domain} shows the results by environment. SIMA outperforms environment-specialized agents overall (67\% average improvement over environment-specialized agent performance), thus demonstrating positive transfer across environments. We statistically quantify this benefit by using a permutation test on the mean difference across the per-task performance of the SIMA agent and the environment-specialized agent within each domain; in every case SIMA significantly outperforms the environment-specialized agent (\(p\)-values on each environment respectively: 0.001, 0.002, 0.036, 0.0002, 0.008, 0.004, and 0.0002). Furthermore, SIMA performs much better than the baselines. SIMA substantially outperforms the no-pretraining baseline overall (permutation test \(p < 0.001\)), thus showing that internet-scale knowledge supports grounded learning---though the magnitude and significance of the benefit varies across the environments (permutation test \(p\)-values respectively 0.0002, 0.14, 0.041, 0.0002, 0.244, 0.052, 0.032). Finally, the no-language ablation performs very poorly (all permutation tests \(p < 0.001\)). Importantly, this demonstrates not only that our agent \emph{is in fact} using language, but also that our evaluation tasks are effectively designed to test this capability, rather than being solvable by simply executing plausible behaviors. 

The zero-shot evaluations are also promising. Even when tested in an environment on which it has not been trained to act the agent demonstrates strong performance on general tasks, though of course it falls short in achieving environment-specific skills. Zero-shot agents are capable of performing generic navigation skills that appear across many games (e.g. ``go down the hill''), and show some more complex abilities like grabbing an object by its color, using the fact that color is consistent across games, and the consistent pattern that most games use left mouse to grab or interact with objects. Importantly, even on the Goat Simulator 3 environment, where the agents have not even received visual finetuning, the zero-shot agent still performs comparably to the environment-specialized one---thus showing transfer is not driven by the visual components alone. Note that even where the numerical performance of the zero-shot and environment-specialized agents is similar, they are generally good at different skills---with the environment-specialized agent performing well on game-specific interactions, but performing more weakly on common skills that are supported across many games, and that the zero-shot agent therefore can execute.

Note that zero-shot performance is especially strong on the WorldLab environment for three reasons. First, the evaluation tasks for this environment contain a relatively larger proportion of domain-general skills, such as recognizing objects by color, because we use them as rapid tests of agent capabilities. Second, this environment uses the same underlying engine and shares some implementation details with the other internal research environments, which may support behavioral transfer despite their varied visual styles, asset libraries, physical mechanics, and environment affordances. Furthermore, environment-specialized agent performance may be slightly weaker on this environment because there is a non-trivial distribution shift from training to test. This is because some of our data comes from earlier versions of the environment with differences in dynamics, and task distributions. Agents trained across multiple environments may be more robust to this distribution shift.

\begin{figure*}
    \centering
    \includegraphics[width=0.7\textwidth]{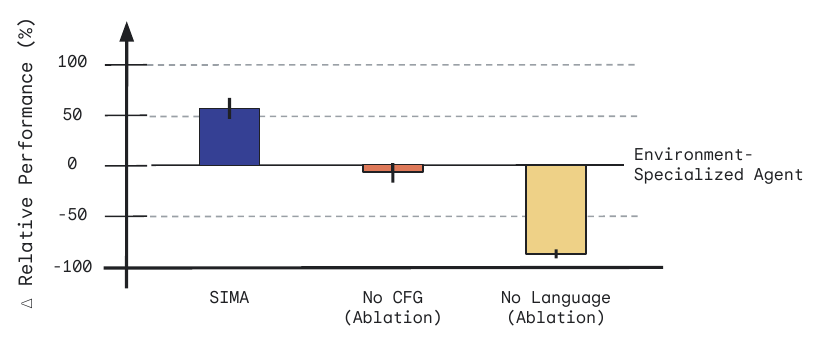}
    \caption{\textbf{Evaluating the Benefit of Classifier-Free Guidance.} Comparing the SIMA agent to an ablation without classifier-free guidance (CFG), CFG substantially improves language conditionality. However, even without CFG, the agent still exhibits language-conditional behavior, outperforming the No Language ablation. Note that this evaluation was performed only on a subset of our research environments: Construction Lab, Playhouse, and WorldLab.}
    \label{fig:cfg_ablation}
\end{figure*}

\paragraph{Classifier-free guidance} Finally, \Cref{fig:cfg_ablation} compares the performance of agents with and without classifier-free guidance \citep[CFG;][]{lifshitz2023steve}, evaluated on a subset of our research environments: Construction Lab, Playhouse, and WorldLab. Without CFG ($\lambda = 0$), the SIMA agent performs noticeably worse. However, the No CFG agent still exhibits a high degree of language conditionality, significantly outperforming the No Language baseline. These results show the benefit of CFG, highlighting the impact that inference-time interventions can have on agent controllability. 

\subsection{Human comparison} \label{sec:results:humans}

To provide an additional baseline comparison, we evaluated our agents against expert human performance on an additional set of tasks from No Man's Sky, which were chosen to test a focused set of skills in a diverse range of settings. These tasks range in difficulty, from simple instructions (\textit{``walk forward''}) to more complex instructions (\textit{``use the analysis visor to identify new animals''}). The humans who performed the tasks were players who participated in our data collection and had experience with the game. We evaluated human performance using the same judges and evaluation setup that was used for our agents; the judges were not told that they were evaluating human performance rather than agents.

\begin{figure*}
    \centering
    \includegraphics[width=0.85\textwidth]{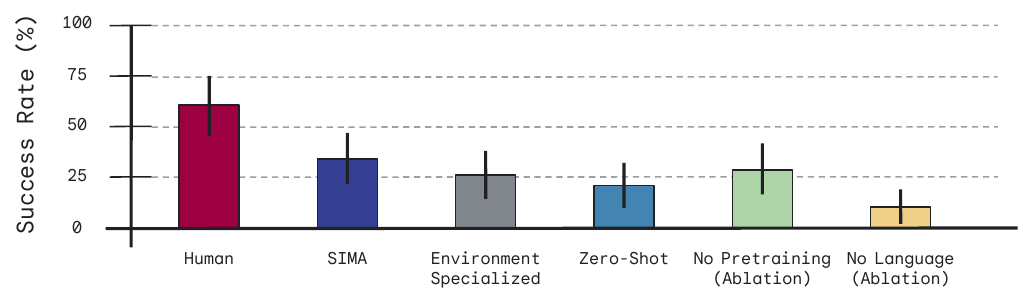}
    \caption{\textbf{Comparison with Human Performance on No Man's Sky.} Evaluating on a subset of tasks from No Man's Sky, human game experts outperform all agents. Yet, humans only achieve 60\% success on this evaluation. This highlights the difficulty of the tasks considered in this project.}
    \label{fig:human_comparison}
\end{figure*}

Results are summarized in \Cref{fig:human_comparison} with error bars denoting parametric 95\%-CIs. The human players achieved a success rate of only 60\% on these tasks, demonstrating the difficulty of the tasks we considered in this project and the stringency of our evaluation criteria. For example, some human failures appear to be due to engaging in unnecessary behaviors before completing the task, like initially opening and interacting with the starship menu when instructed to \textit{``recharge the mining beam,''} or entering analysis mode after scanning when told to \textit{``mine oxygen.''} Despite these challenging evaluations, the SIMA agent achieved non-trivial performance (34\% success), far exceeding that of the No Language baseline (11\% success), for example. We note that 100\% success may not necessarily be achievable, due to disagreement between human judges on more ambiguous tasks. Nevertheless, there is still considerable progress needed to match human performance. This underscores the utility of the entire SIMA setup for providing a challenging, yet informative, metric for assessing grounded language interactions in embodied agents.

\section{Looking ahead}
\label{sec:discussion}

SIMA is a work in progress. In this tech report, we have described our goal and philosophy, and presented some preliminary results showing our agent's ability to ground language instructions in behavior across a variety of rich 3D environments. We see notable performance and early signs of transfer across environments, as well as zero-shot transfer of basic skills to held-out environments. Still, many skills and tasks remain out of reach. In our future work, we aim to \textbf{a)} scale to more environments and datasets by continuing to expand our portfolio of games, environments, and datasets; \textbf{b)} increase the robustness and controllability of agents; \textbf{c)} leverage increasingly high-quality pretrained models \citep{geminiteam2023gemini}; and \textbf{d)} develop more comprehensive and carefully controlled evaluations.

We believe that by doing so, we will make SIMA an ideal platform for doing cutting-edge research on grounding language and pretrained models safely in complex environments, thereby helping to tackle a fundamental challenge of AGI. Our research also has the potential to enrich the learning experiences and deployment environments of future foundation models; one of our goals is to ground the abstract capabilities of large language models in embodied environments. We hope that SIMA will help us learn how to overcome the fundamental challenge of linking language to perception and action at scale, and we are excited to share more details about our research in the future.

\section*{Acknowledgements}

We thank the following games developers for partnering with us on this project: Coffee Stain, Foulball Hangover, Hello Games, Keen Software House, Rubberband Games, Saber Interactive / Tuxedo Labs, and Strange Loop Games.
We also thank \cite{bica2024improving} for their assistance in incorporating SPARC into the SIMA agent as well as \cite{zolna2024gats} and Scott Reed for their assistance in incorporating Phenaki into the SIMA agent.
We thank Matthew McGill, Nicholas Roy, Avraham Ruderman, Daniel Tanis, and  Frank Perbet for their assistance with research environment task development. We thank Alistair Muldal for assistance with data and infrastructure from prior efforts. We also thank Timothy Lillicrap for early input into the SIMA concept and insights from prior efforts.
We thank Tom Ward, Joe Stanton, David Barker, and George Thomas for their infrastructure and support for running game binaries on Google Cloud infrastructure.

Finally, we thank our team of participants who generated gameplay and language annotation data, as well as performed human evaluations of our agents, without whom this work would not have been possible.

\bibliography{main}

\begin{thebibliography}{92}
\providecommand{\natexlab}[1]{#1}
\providecommand{\url}[1]{\texttt{#1}}
\expandafter\ifx\csname urlstyle\endcsname\relax
  \providecommand{\doi}[1]{doi: #1}\else
  \providecommand{\doi}{doi: \begingroup \urlstyle{rm}\Url}\fi

\bibitem[Abramson et~al.(2020)Abramson, Ahuja, Barr, Brussee, Carnevale,
  Cassin, Chhaparia, Clark, Damoc, Dudzik, et~al.]{abramson2020imitating}
Josh Abramson, Arun Ahuja, Iain Barr, Arthur Brussee, Federico Carnevale, Mary
  Cassin, Rachita Chhaparia, Stephen Clark, Bogdan Damoc, Andrew Dudzik, et~al.
\newblock {Imitating Interactive Intelligence}.
\newblock \emph{arXiv preprint arXiv:2012.05672}, 2020.

\bibitem[Abramson et~al.(2022{\natexlab{a}})Abramson, Ahuja, Carnevale,
  Georgiev, Goldin, Hung, Landon, Lhotka, Lillicrap, Muldal,
  et~al.]{abramson2022improving}
Josh Abramson, Arun Ahuja, Federico Carnevale, Petko Georgiev, Alex Goldin,
  Alden Hung, Jessica Landon, Jirka Lhotka, Timothy Lillicrap, Alistair Muldal,
  et~al.
\newblock {Improving Multimodal Interactive Agents with Reinforcement Learning
  from Human Feedback}.
\newblock \emph{arXiv preprint arXiv:2211.11602}, 2022{\natexlab{a}}.

\bibitem[Abramson et~al.(2022{\natexlab{b}})Abramson, Ahuja, Carnevale,
  Georgiev, Goldin, Hung, Landon, Lillicrap, Muldal, Richards,
  et~al.]{abramson2022evaluating}
Josh Abramson, Arun Ahuja, Federico Carnevale, Petko Georgiev, Alex Goldin,
  Alden Hung, Jessica Landon, Timothy Lillicrap, Alistair Muldal, Blake
  Richards, et~al.
\newblock {Evaluating Multimodal Interactive Agents}.
\newblock \emph{arXiv preprint arXiv:2205.13274}, 2022{\natexlab{b}}.

\bibitem[{Adaptive Agent Team} et~al.(2023){Adaptive Agent Team}, Bauer,
  Baumli, Baveja, Behbahani, Bhoopchand, Bradley-Schmieg, Chang, Clay,
  Collister, et~al.]{team2023human}
{Adaptive Agent Team}, Jakob Bauer, Kate Baumli, Satinder Baveja, Feryal
  Behbahani, Avishkar Bhoopchand, Nathalie Bradley-Schmieg, Michael Chang,
  Natalie Clay, Adrian Collister, et~al.
\newblock {Human-Timescale Adaptation in an Open-Ended Task Space}.
\newblock In \emph{International Conference on Machine Learning}, 2023.

\bibitem[Ajay et~al.(2023)Ajay, Han, Du, Li, Gupta, Jaakkola, Tenenbaum,
  Kaelbling, Srivastava, and Agrawal]{ajay2023compositional}
Anurag Ajay, Seungwook Han, Yilun Du, Shuang Li, Abhi Gupta, Tommi Jaakkola,
  Josh Tenenbaum, Leslie Kaelbling, Akash Srivastava, and Pulkit Agrawal.
\newblock {Compositional Foundation Models for Hierarchical Planning}.
\newblock In \emph{Advances in Neural Information Processing Systems}, 2023.

\bibitem[Albrecht et~al.(2022)Albrecht, Fetterman, Fogelman, Kitanidis,
  Wr{\'o}blewski, Seo, Rosenthal, Knutins, Polizzi, Simon,
  et~al.]{albrecht2022avalon}
Joshua Albrecht, Abraham Fetterman, Bryden Fogelman, Ellie Kitanidis, Bartosz
  Wr{\'o}blewski, Nicole Seo, Michael Rosenthal, Maksis Knutins, Zack Polizzi,
  James Simon, et~al.
\newblock {Avalon: A Benchmark for RL Generalization Using Procedurally
  Generated Worlds}.
\newblock In \emph{Advances in Neural Information Processing Systems}, 2022.

\bibitem[Anil et~al.(2023)Anil, Dai, Firat, Johnson, Lepikhin, Passos, Shakeri,
  Taropa, Bailey, Chen, et~al.]{anil2023palm}
Rohan Anil, Andrew~M Dai, Orhan Firat, Melvin Johnson, Dmitry Lepikhin,
  Alexandre Passos, Siamak Shakeri, Emanuel Taropa, Paige Bailey, Zhifeng Chen,
  et~al.
\newblock {PaLM 2 Technical Report}.
\newblock \emph{arXiv preprint arXiv:2305.10403}, 2023.

\bibitem[Baker et~al.(2022)Baker, Akkaya, Zhokov, Huizinga, Tang, Ecoffet,
  Houghton, Sampedro, and Clune]{baker2022video}
Bowen Baker, Ilge Akkaya, Peter Zhokov, Joost Huizinga, Jie Tang, Adrien
  Ecoffet, Brandon Houghton, Raul Sampedro, and Jeff Clune.
\newblock {Video PreTraining (VPT): Learning to Act by Watching Unlabeled
  Online Videos}.
\newblock In \emph{Advances in Neural Information Processing Systems}, 2022.

\bibitem[Bellemare et~al.(2013)Bellemare, Naddaf, Veness, and
  Bowling]{bellemare2013arcade}
Marc~G Bellemare, Yavar Naddaf, Joel Veness, and Michael Bowling.
\newblock {The Arcade Learning Environment: An Evaluation Platform for General
  Agents}.
\newblock \emph{Journal of Artificial Intelligence Research}, 47:\penalty0
  253--279, 2013.

\bibitem[Berner et~al.(2019)Berner, Brockman, Chan, Cheung, D{\k{e}}biak,
  Dennison, Farhi, Fischer, Hashme, Hesse, et~al.]{berner2019dota}
Christopher Berner, Greg Brockman, Brooke Chan, Vicki Cheung, Przemys{\l}aw
  D{\k{e}}biak, Christy Dennison, David Farhi, Quirin Fischer, Shariq Hashme,
  Chris Hesse, et~al.
\newblock {Dota 2 with Large Scale Deep Reinforcement Learning}.
\newblock \emph{arXiv preprint arXiv:1912.06680}, 2019.

\bibitem[Bica et~al.(2024)Bica, Ilić, Bauer, Erdogan, Bošnjak, Kaplanis,
  Gritsenko, Minderer, Blundell, Pascanu, and Mitrović]{bica2024improving}
Ioana Bica, Anastasija Ilić, Matthias Bauer, Goker Erdogan, Matko Bošnjak,
  Christos Kaplanis, Alexey~A. Gritsenko, Matthias Minderer, Charles Blundell,
  Razvan Pascanu, and Jovana Mitrović.
\newblock {Improving fine-grained understanding in image-text pre-training}.
\newblock \emph{arXiv preprint arXiv:2401.09865}, 2024.

\bibitem[Bratko et~al.(1995)Bratko, Urbančič, and
  Sammut]{bratko1995behavioural}
Ivan Bratko, Tanja Urbančič, and Claude Sammut.
\newblock {Behavioural Cloning: Phenomena, Results and Problems}.
\newblock \emph{IFAC Proceedings Volumes}, 28\penalty0 (21):\penalty0 143--149,
  1995.

\bibitem[Brohan et~al.(2022)Brohan, Brown, Carbajal, Chebotar, Dabis, Finn,
  Gopalakrishnan, Hausman, Herzog, Hsu, et~al.]{brohan2022rt1}
Anthony Brohan, Noah Brown, Justice Carbajal, Yevgen Chebotar, Joseph Dabis,
  Chelsea Finn, Keerthana Gopalakrishnan, Karol Hausman, Alex Herzog, Jasmine
  Hsu, et~al.
\newblock {RT-1: Robotics Transformer for Real-World Control at Scale}.
\newblock \emph{arXiv preprint arXiv:2212.06817}, 2022.

\bibitem[Brohan et~al.(2023{\natexlab{a}})Brohan, Brown, Carbajal, Chebotar,
  Chen, Choromanski, Ding, Driess, Dubey, Finn, et~al.]{brohan2023rt2}
Anthony Brohan, Noah Brown, Justice Carbajal, Yevgen Chebotar, Xi~Chen,
  Krzysztof Choromanski, Tianli Ding, Danny Driess, Avinava Dubey, Chelsea
  Finn, et~al.
\newblock {RT-2: Vision-Language-Action Models Transfer Web Knowledge to
  Robotic Control}.
\newblock \emph{arXiv preprint arXiv:2307.15818}, 2023{\natexlab{a}}.

\bibitem[Brohan et~al.(2023{\natexlab{b}})Brohan, Chebotar, Finn, Hausman,
  Herzog, Ho, Ibarz, Irpan, Jang, Julian, et~al.]{brohan2023can}
Anthony Brohan, Yevgen Chebotar, Chelsea Finn, Karol Hausman, Alexander Herzog,
  Daniel Ho, Julian Ibarz, Alex Irpan, Eric Jang, Ryan Julian, et~al.
\newblock {Do As I Can, Not As I Say: Grounding Language in Robotic
  Affordances}.
\newblock In \emph{Conference on Robot Learning}, 2023{\natexlab{b}}.

\bibitem[Brown et~al.(2020)Brown, Mann, Ryder, Subbiah, Kaplan, Dhariwal,
  Neelakantan, Shyam, Sastry, Askell, et~al.]{brown2020language}
Tom Brown, Benjamin Mann, Nick Ryder, Melanie Subbiah, Jared~D Kaplan, Prafulla
  Dhariwal, Arvind Neelakantan, Pranav Shyam, Girish Sastry, Amanda Askell,
  et~al.
\newblock {Language Models are Few-Shot Learners}.
\newblock In \emph{Advances in Neural Information Processing Systems}, 2020.

\bibitem[Colas et~al.(2020)Colas, Karch, Lair, Dussoux, Moulin-Frier, Dominey,
  and Oudeyer]{colas2020language}
C{\'e}dric Colas, Tristan Karch, Nicolas Lair, Jean-Michel Dussoux, Cl{\'e}ment
  Moulin-Frier, Peter Dominey, and Pierre-Yves Oudeyer.
\newblock {Language as a Cognitive Tool to Imagine Goals in Curiosity-Driven
  Exploration}.
\newblock In \emph{Advances in Neural Information Processing Systems}, 2020.

\bibitem[Colas et~al.(2022)Colas, Karch, Moulin-Frier, and
  Oudeyer]{colas2022language}
C{\'e}dric Colas, Tristan Karch, Cl{\'e}ment Moulin-Frier, and Pierre-Yves
  Oudeyer.
\newblock {Language and culture internalization for human-like autotelic AI}.
\newblock \emph{Nature Machine Intelligence}, 4\penalty0 (12):\penalty0
  1068--1076, 2022.

\bibitem[Coumans and Bai(2016)]{coumans2016}
Erwin Coumans and Yunfei Bai.
\newblock {PyBullet, a Python module for physics simulation for games, robotics
  and machine learning}.
\newblock \url{http://pybullet.org}, 2016.

\bibitem[Dai et~al.(2019)Dai, Yang, Yang, Carbonell, Le, and
  Salakhutdinov]{dai2019transformer}
Zihang Dai, Zhilin Yang, Yiming Yang, Jaime~G Carbonell, Quoc Le, and Ruslan
  Salakhutdinov.
\newblock {Transformer-XL: Attentive Language Models beyond a Fixed-Length
  Context}.
\newblock In \emph{Association for Computational Linguistics}, 2019.

\bibitem[{DeepMind Interactive Agents Team} et~al.(2021){DeepMind Interactive
  Agents Team}, Abramson, Ahuja, Brussee, Carnevale, Cassin, Fischer, Georgiev,
  Goldin, Gupta, et~al.]{team2021creating}
{DeepMind Interactive Agents Team}, Josh Abramson, Arun Ahuja, Arthur Brussee,
  Federico Carnevale, Mary Cassin, Felix Fischer, Petko Georgiev, Alex Goldin,
  Mansi Gupta, et~al.
\newblock {Creating Multimodal Interactive Agents with Imitation and
  Self-Supervised Learning}.
\newblock \emph{arXiv preprint arXiv:2112.03763}, 2021.

\bibitem[Deitke et~al.(2022)Deitke, VanderBilt, Herrasti, Weihs, Ehsani,
  Salvador, Han, Kolve, Kembhavi, and Mottaghi]{deitke2022procthor}
Matt Deitke, Eli VanderBilt, Alvaro Herrasti, Luca Weihs, Kiana Ehsani, Jordi
  Salvador, Winson Han, Eric Kolve, Aniruddha Kembhavi, and Roozbeh Mottaghi.
\newblock {ProcTHOR: Large-Scale Embodied AI Using Procedural Generation}.
\newblock In \emph{Advances in Neural Information Processing Systems}, 2022.

\bibitem[Dosovitskiy et~al.(2017)Dosovitskiy, Ros, Codevilla, Lopez, and
  Koltun]{dosovitskiy2017carla}
Alexey Dosovitskiy, German Ros, Felipe Codevilla, Antonio Lopez, and Vladlen
  Koltun.
\newblock {CARLA: An Open Urban Driving Simulator}.
\newblock In \emph{Conference on Robot Learning}, 2017.

\bibitem[Driess et~al.(2023)Driess, Xia, Sajjadi, Lynch, Chowdhery, Ichter,
  Wahid, Tompson, Vuong, Yu, et~al.]{driess2023palm}
Danny Driess, Fei Xia, Mehdi~SM Sajjadi, Corey Lynch, Aakanksha Chowdhery,
  Brian Ichter, Ayzaan Wahid, Jonathan Tompson, Quan Vuong, Tianhe Yu, et~al.
\newblock {PaLM-E: An Embodied Multimodal Language Model}.
\newblock \emph{arXiv preprint arXiv:2303.03378}, 2023.

\bibitem[Durante et~al.(2024)Durante, Sarkar, Gong, Taori, Noda, Tang, Adeli,
  Lakshmikanth, Schulman, Milstein, et~al.]{durante2024interactive}
Zane Durante, Bidipta Sarkar, Ran Gong, Rohan Taori, Yusuke Noda, Paul Tang,
  Ehsan Adeli, Shrinidhi~Kowshika Lakshmikanth, Kevin Schulman, Arnold
  Milstein, et~al.
\newblock {An Interactive Agent Foundation Model}.
\newblock \emph{arXiv preprint arXiv:2402.05929}, 2024.

\bibitem[Espeholt et~al.(2018)Espeholt, Soyer, Munos, Simonyan, Mnih, Ward,
  Doron, Firoiu, Harley, Dunning, et~al.]{espeholt2018impala}
Lasse Espeholt, Hubert Soyer, Remi Munos, Karen Simonyan, Vlad Mnih, Tom Ward,
  Yotam Doron, Vlad Firoiu, Tim Harley, Iain Dunning, et~al.
\newblock {IMPALA: Scalable Distributed Deep-RL with Importance Weighted
  Actor-Learner Architectures}.
\newblock In \emph{International Conference on Machine Learning}, 2018.

\bibitem[Fan et~al.(2022)Fan, Wang, Jiang, Mandlekar, Yang, Zhu, Tang, Huang,
  Zhu, and Anandkumar]{fan2022minedojo}
Linxi Fan, Guanzhi Wang, Yunfan Jiang, Ajay Mandlekar, Yuncong Yang, Haoyi Zhu,
  Andrew Tang, De-An Huang, Yuke Zhu, and Anima Anandkumar.
\newblock {MineDojo: Building Open-Ended Embodied Agents with Internet-Scale
  Knowledge}.
\newblock In \emph{Advances in Neural Information Processing Systems}, 2022.

\bibitem[{Gemini Team} et~al.(2023){Gemini Team}, Anil, Borgeaud, Wu, Alayrac,
  Yu, Soricut, Schalkwyk, Dai, Hauth, et~al.]{geminiteam2023gemini}
{Gemini Team}, Rohan Anil, Sebastian Borgeaud, Yonghui Wu, Jean-Baptiste
  Alayrac, Jiahui Yu, Radu Soricut, Johan Schalkwyk, Andrew~M Dai, Anja Hauth,
  et~al.
\newblock {Gemini: A Family of Highly Capable Multimodal Models}.
\newblock \emph{arXiv preprint arXiv:2312.11805}, 2023.

\bibitem[Gregor et~al.(2019)Gregor, Jimenez~Rezende, Besse, Wu, Merzic, and
  van~den Oord]{gregor2019shaping}
Karol Gregor, Danilo Jimenez~Rezende, Frederic Besse, Yan Wu, Hamza Merzic, and
  Aaron van~den Oord.
\newblock {Shaping Belief States with Generative Environment Models for RL}.
\newblock In \emph{Advances in Neural Information Processing Systems}, 2019.

\bibitem[Gulcehre et~al.(2019)Gulcehre, Le~Paine, Shahriari, Denil, Hoffman,
  Soyer, Tanburn, Kapturowski, Rabinowitz, Williams, et~al.]{paine2019making}
Caglar Gulcehre, Tom Le~Paine, Bobak Shahriari, Misha Denil, Matt Hoffman,
  Hubert Soyer, Richard Tanburn, Steven Kapturowski, Neil Rabinowitz, Duncan
  Williams, et~al.
\newblock {Making Efficient Use of Demonstrations to Solve Hard Exploration
  Problems}.
\newblock In \emph{International Conference on Learning Representations}, 2019.

\bibitem[Guss et~al.(2019)Guss, Houghton, Topin, Wang, Codel, Veloso, and
  Salakhutdinov]{guss2019minerl}
William~H Guss, Brandon Houghton, Nicholay Topin, Phillip Wang, Cayden Codel,
  Manuela Veloso, and Ruslan Salakhutdinov.
\newblock {MineRL: A Large-Scale Dataset of Minecraft Demonstrations}.
\newblock In \emph{International Joint Conference on Artificial Intelligence},
  2019.

\bibitem[Ha and Schmidhuber(2018)]{ha2018world}
David Ha and J{\"u}rgen Schmidhuber.
\newblock {Recurrent World Models Facilitate Policy Evolution}.
\newblock In \emph{Advances in Neural Information Processing Systems}, 2018.

\bibitem[Hafner et~al.(2020)Hafner, Lillicrap, Norouzi, and
  Ba]{hafner2020mastering}
Danijar Hafner, Timothy~P Lillicrap, Mohammad Norouzi, and Jimmy Ba.
\newblock {Mastering Atari with Discrete World Models}.
\newblock In \emph{International Conference on Learning Representations}, 2020.

\bibitem[Hafner et~al.(2023)Hafner, Pasukonis, Ba, and
  Lillicrap]{hafner2023mastering}
Danijar Hafner, Jurgis Pasukonis, Jimmy Ba, and Timothy Lillicrap.
\newblock {Mastering Diverse Domains through World Models}.
\newblock \emph{arXiv preprint arXiv:2301.04104}, 2023.

\bibitem[Harnad(1990)]{harnad1990symbol}
Stevan Harnad.
\newblock {The Symbol Grounding Problem}.
\newblock \emph{Physica D: Nonlinear Phenomena}, 42\penalty0 (1-3):\penalty0
  335--346, 1990.

\bibitem[Hermann et~al.(2017)Hermann, Hill, Green, Wang, Faulkner, Soyer,
  Szepesvari, Czarnecki, Jaderberg, Teplyashin, et~al.]{hermann2017grounded}
Karl~Moritz Hermann, Felix Hill, Simon Green, Fumin Wang, Ryan Faulkner, Hubert
  Soyer, David Szepesvari, Wojciech~Marian Czarnecki, Max Jaderberg, Denis
  Teplyashin, et~al.
\newblock {Grounded Language Learning in a Simulated 3D World}.
\newblock \emph{arXiv preprint arXiv:1706.06551}, 2017.

\bibitem[Hill et~al.(2019)Hill, Lampinen, Schneider, Clark, Botvinick,
  McClelland, and Santoro]{hill2019environmental}
Felix Hill, Andrew Lampinen, Rosalia Schneider, Stephen Clark, Matthew
  Botvinick, James~L McClelland, and Adam Santoro.
\newblock {Environmental drivers of systematicity and generalization in a
  situated agent}.
\newblock In \emph{International Conference on Learning Representations}, 2019.

\bibitem[Hill et~al.(2020)Hill, Tieleman, von Glehn, Wong, Merzic, and
  Clark]{hill2020grounded}
Felix Hill, Olivier Tieleman, Tamara von Glehn, Nathaniel Wong, Hamza Merzic,
  and Stephen Clark.
\newblock {Grounded Language Learning Fast and Slow}.
\newblock In \emph{International Conference on Learning Representations}, 2020.

\bibitem[Ho and Salimans(2022)]{ho2022classifier}
Jonathan Ho and Tim Salimans.
\newblock {Classifier-Free Diffusion Guidance}.
\newblock \emph{arXiv preprint arXiv:2207.12598}, 2022.

\bibitem[H{\"o}fer et~al.(2021)H{\"o}fer, Bekris, Handa, Gamboa, Mozifian,
  Golemo, Atkeson, Fox, Goldberg, Leonard, et~al.]{hofer2021sim2real}
Sebastian H{\"o}fer, Kostas Bekris, Ankur Handa, Juan~Camilo Gamboa, Melissa
  Mozifian, Florian Golemo, Chris Atkeson, Dieter Fox, Ken Goldberg, John
  Leonard, et~al.
\newblock {Sim2Real in Robotics and Automation: Applications and Challenges}.
\newblock \emph{IEEE Transactions on Automation Science and Engineering},
  18\penalty0 (2):\penalty0 398--400, 2021.

\bibitem[Hoffmann et~al.(2022)Hoffmann, Borgeaud, Mensch, Buchatskaya, Cai,
  Rutherford, Casas, Hendricks, Welbl, Clark, et~al.]{hoffmann2022training}
Jordan Hoffmann, Sebastian Borgeaud, Arthur Mensch, Elena Buchatskaya, Trevor
  Cai, Eliza Rutherford, Diego de~Las Casas, Lisa~Anne Hendricks, Johannes
  Welbl, Aidan Clark, et~al.
\newblock {Training Compute-Optimal Large Language Models}.
\newblock \emph{arXiv preprint arXiv:2203.15556}, 2022.

\bibitem[Hu and Clune(2023)]{hu2023thought}
Shengran Hu and Jeff Clune.
\newblock {Thought Cloning: Learning to Think while Acting by Imitating Human
  Thinking}.
\newblock \emph{arXiv preprint arXiv:2306.00323}, 2023.

\bibitem[Hu et~al.(2023)Hu, Lin, Zhang, Yi, and Gao]{hu2023look}
Yingdong Hu, Fanqi Lin, Tong Zhang, Li~Yi, and Yang Gao.
\newblock {Look Before You Leap: Unveiling the Power of GPT-4V in Robotic
  Vision-Language Planning}.
\newblock \emph{arXiv preprint arXiv:2311.17842}, 2023.

\bibitem[Huang et~al.(2023)Huang, Yong, Ma, Linghu, Li, Wang, Li, Zhu, Jia, and
  Huang]{huang2023embodied}
Jiangyong Huang, Silong Yong, Xiaojian Ma, Xiongkun Linghu, Puhao Li, Yan Wang,
  Qing Li, Song-Chun Zhu, Baoxiong Jia, and Siyuan Huang.
\newblock {An Embodied Generalist Agent in 3D World}.
\newblock \emph{arXiv preprint arXiv:2311.12871}, 2023.

\bibitem[Huang et~al.(2022)Huang, Abbeel, Pathak, and
  Mordatch]{huang2022language}
Wenlong Huang, Pieter Abbeel, Deepak Pathak, and Igor Mordatch.
\newblock {Language Models as Zero-Shot Planners: Extracting Actionable
  Knowledge for Embodied Agents}.
\newblock In \emph{International Conference on Machine Learning}, 2022.

\bibitem[Humphreys et~al.(2022)Humphreys, Raposo, Pohlen, Thornton, Chhaparia,
  Muldal, Abramson, Georgiev, Santoro, and Lillicrap]{humphreys2022data}
Peter~C Humphreys, David Raposo, Tobias Pohlen, Gregory Thornton, Rachita
  Chhaparia, Alistair Muldal, Josh Abramson, Petko Georgiev, Adam Santoro, and
  Timothy Lillicrap.
\newblock {A data-driven approach for learning to control computers}.
\newblock In \emph{International Conference on Machine Learning}, 2022.

\bibitem[Jiang et~al.(2019)Jiang, Gu, Murphy, and Finn]{jiang2019language}
Yiding Jiang, Shixiang~Shane Gu, Kevin~P Murphy, and Chelsea Finn.
\newblock {Language as an Abstraction for Hierarchical Deep Reinforcement
  Learning}.
\newblock In \emph{Advances in Neural Information Processing Systems}, 2019.

\bibitem[Johnson et~al.(2016)Johnson, Hofmann, Hutton, and
  Bignell]{johnson2016malmo}
Matthew Johnson, Katja Hofmann, Tim Hutton, and David Bignell.
\newblock {The Malmo Platform for Artificial Intelligence Experimentation}.
\newblock In \emph{International Joint Conference on Artificial Intelligence},
  2016.

\bibitem[Kim et~al.(2023)Kim, Baldi, and McAleer]{kim2023language}
Geunwoo Kim, Pierre Baldi, and Stephen McAleer.
\newblock {Language Models can Solve Computer Tasks}.
\newblock In \emph{Advances in Neural Information Processing Systems}, 2023.

\bibitem[Koh et~al.(2024)Koh, Lo, Jang, Duvvur, Lim, Huang, Neubig, Zhou,
  Salakhutdinov, and Fried]{koh2024visualwebarena}
Jing~Yu Koh, Robert Lo, Lawrence Jang, Vikram Duvvur, Ming~Chong Lim, Po-Yu
  Huang, Graham Neubig, Shuyan Zhou, Ruslan Salakhutdinov, and Daniel Fried.
\newblock {VisualWebArena: Evaluating Multimodal Agents on Realistic Visual Web
  Tasks}.
\newblock \emph{arXiv preprint arXiv:2401.13649}, 2024.

\bibitem[Kolve et~al.(2017)Kolve, Mottaghi, Han, VanderBilt, Weihs, Herrasti,
  Deitke, Ehsani, Gordon, Zhu, et~al.]{kolve2017ai2}
Eric Kolve, Roozbeh Mottaghi, Winson Han, Eli VanderBilt, Luca Weihs, Alvaro
  Herrasti, Matt Deitke, Kiana Ehsani, Daniel Gordon, Yuke Zhu, et~al.
\newblock {AI2-THOR: An Interactive 3D Environment for Visual AI}.
\newblock \emph{arXiv preprint arXiv:1712.05474}, 2017.

\bibitem[Kumar et~al.(2022)Kumar, Correa, Dasgupta, Marjieh, Hu, Hawkins,
  Cohen, Narasimhan, Griffiths, et~al.]{kumar2022using}
Sreejan Kumar, Carlos~G Correa, Ishita Dasgupta, Raja Marjieh, Michael~Y Hu,
  Robert Hawkins, Jonathan~D Cohen, Karthik Narasimhan, Tom Griffiths, et~al.
\newblock {Using Natural Language and Program Abstractions to Instill Human
  Inductive Biases in Machines}.
\newblock In \emph{Advances in Neural Information Processing Systems}, 2022.

\bibitem[Lampinen et~al.(2022)Lampinen, Roy, Dasgupta, Chan, Tam, Mcclelland,
  Yan, Santoro, Rabinowitz, Wang, et~al.]{lampinen2022tell}
Andrew~K Lampinen, Nicholas Roy, Ishita Dasgupta, Stephanie~CY Chan, Allison
  Tam, James Mcclelland, Chen Yan, Adam Santoro, Neil~C Rabinowitz, Jane Wang,
  et~al.
\newblock {Tell me why! Explanations support learning relational and causal
  structure}.
\newblock In \emph{International Conference on Machine Learning}, 2022.

\bibitem[Li et~al.(2022)Li, Choi, Chung, Kushman, Schrittwieser, Leblond,
  Eccles, Keeling, Gimeno, Dal~Lago, et~al.]{li2022competition}
Yujia Li, David Choi, Junyoung Chung, Nate Kushman, Julian Schrittwieser,
  R{\'e}mi Leblond, Tom Eccles, James Keeling, Felix Gimeno, Agustin Dal~Lago,
  et~al.
\newblock {Competition-Level Code Generation with AlphaCode}.
\newblock \emph{Science}, 378\penalty0 (6624):\penalty0 1092--1097, 2022.

\bibitem[Lifshitz et~al.(2023)Lifshitz, Paster, Chan, Ba, and
  McIlraith]{lifshitz2023steve}
Shalev Lifshitz, Keiran Paster, Harris Chan, Jimmy Ba, and Sheila McIlraith.
\newblock {STEVE-1: A Generative Model for Text-to-Behavior in Minecraft}.
\newblock \emph{arXiv preprint arXiv:2306.00937}, 2023.

\bibitem[Makoviychuk et~al.(2021)Makoviychuk, Wawrzyniak, Guo, Lu, Storey,
  Macklin, Hoeller, Rudin, Allshire, Handa, et~al.]{makoviychuk2021isaac}
Viktor Makoviychuk, Lukasz Wawrzyniak, Yunrong Guo, Michelle Lu, Kier Storey,
  Miles Macklin, David Hoeller, Nikita Rudin, Arthur Allshire, Ankur Handa,
  et~al.
\newblock {Isaac Gym: High Performance GPU Based Physics Simulation For Robot
  Learning}.
\newblock In \emph{Advances in Neural Information Processing Systems}, 2021.

\bibitem[McClelland et~al.(2020)McClelland, Hill, Rudolph, Baldridge, and
  Sch{\"u}tze]{mcclelland2020placing}
James~L McClelland, Felix Hill, Maja Rudolph, Jason Baldridge, and Hinrich
  Sch{\"u}tze.
\newblock {Placing language in an integrated understanding system: Next steps
  toward human-level performance in neural language models}.
\newblock \emph{Proceedings of the National Academy of Sciences}, 117\penalty0
  (42):\penalty0 25966--25974, 2020.

\bibitem[Mnih et~al.(2015)Mnih, Kavukcuoglu, Silver, Rusu, Veness, Bellemare,
  Graves, Riedmiller, Fidjeland, Ostrovski, et~al.]{mnih2015human}
Volodymyr Mnih, Koray Kavukcuoglu, David Silver, Andrei~A Rusu, Joel Veness,
  Marc~G Bellemare, Alex Graves, Martin Riedmiller, Andreas~K Fidjeland, Georg
  Ostrovski, et~al.
\newblock {Human-level control through deep reinforcement learning}.
\newblock \emph{Nature}, 518\penalty0 (7540):\penalty0 529--533, 2015.

\bibitem[Moravec(1988)]{moravec1988mind}
Hans Moravec.
\newblock \emph{{Mind Children: The Future of Robot and Human Intelligence}}.
\newblock Harvard University Press, 1988.

\bibitem[Mu et~al.(2022)Mu, Zhong, Raileanu, Jiang, Goodman, Rockt{\"a}schel,
  and Grefenstette]{mu2022improving}
Jesse Mu, Victor Zhong, Roberta Raileanu, Minqi Jiang, Noah Goodman, Tim
  Rockt{\"a}schel, and Edward Grefenstette.
\newblock {Improving Intrinsic Exploration with Language Abstractions}.
\newblock In \emph{Advances in Neural Information Processing Systems}, 2022.

\bibitem[Nottingham et~al.(2023)Nottingham, Ammanabrolu, Suhr, Choi,
  Hajishirzi, Singh, and Fox]{nottingham2023embodied}
Kolby Nottingham, Prithviraj Ammanabrolu, Alane Suhr, Yejin Choi, Hannaneh
  Hajishirzi, Sameer Singh, and Roy Fox.
\newblock {Do Embodied Agents Dream of Pixelated Sheep: Embodied Decision
  Making using Language Guided World Modelling}.
\newblock \emph{arXiv preprint arXiv:2301.12050}, 2023.

\bibitem[{Open Ended Learning Team} et~al.(2021){Open Ended Learning Team},
  Stooke, Mahajan, Barros, Deck, Bauer, Sygnowski, Trebacz, Jaderberg, Mathieu,
  et~al.]{team2021open}
{Open Ended Learning Team}, Adam Stooke, Anuj Mahajan, Catarina Barros, Charlie
  Deck, Jakob Bauer, Jakub Sygnowski, Maja Trebacz, Max Jaderberg, Michael
  Mathieu, et~al.
\newblock {Open-Ended Learning Leads to Generally Capable Agents}.
\newblock \emph{arXiv preprint arXiv:2107.12808}, 2021.

\bibitem[{OpenAI}(2023)]{openai2023gpt4}
{OpenAI}.
\newblock {GPT-4 Technical Report}.
\newblock \emph{arXiv preprint arXiv:2303.08774}, 2023.

\bibitem[Padalkar et~al.(2023)Padalkar, Pooley, Jain, Bewley, Herzog, Irpan,
  Khazatsky, Rai, Singh, Brohan, et~al.]{padalkar2023open}
Abhishek Padalkar, Acorn Pooley, Ajinkya Jain, Alex Bewley, Alex Herzog, Alex
  Irpan, Alexander Khazatsky, Anant Rai, Anikait Singh, Anthony Brohan, et~al.
\newblock {Open X-Embodiment: Robotic Learning Datasets and RT-X Models}.
\newblock \emph{arXiv preprint arXiv:2310.08864}, 2023.

\bibitem[Pearce and Zhu(2022)]{pearce2022counter}
Tim Pearce and Jun Zhu.
\newblock {Counter-Strike Deathmatch with Large-Scale Behavioural Cloning}.
\newblock In \emph{IEEE Conference on Games}, 2022.

\bibitem[Puig et~al.(2018)Puig, Ra, Boben, Li, Wang, Fidler, and
  Torralba]{puig2018virtualhome}
Xavier Puig, Kevin Ra, Marko Boben, Jiaman Li, Tingwu Wang, Sanja Fidler, and
  Antonio Torralba.
\newblock {VirtualHome: Simulating Household Activities via Programs}.
\newblock In \emph{Computer Vision and Pattern Recognition}, 2018.

\bibitem[Puig et~al.(2023)Puig, Undersander, Szot, Cote, Yang, Partsey, Desai,
  Clegg, Hlavac, Min, Vondruš, Gervet, Berges, Turner, Maksymets, Kira,
  Kalakrishnan, Malik, Chaplot, Jain, Batra, Rai, and
  Mottaghi]{puig2023habitat}
Xavier Puig, Eric Undersander, Andrew Szot, Mikael~Dallaire Cote, Tsung-Yen
  Yang, Ruslan Partsey, Ruta Desai, Alexander~William Clegg, Michal Hlavac,
  So~Yeon Min, Vladimír Vondruš, Theophile Gervet, Vincent-Pierre Berges,
  John~M. Turner, Oleksandr Maksymets, Zsolt Kira, Mrinal Kalakrishnan,
  Jitendra Malik, Devendra~Singh Chaplot, Unnat Jain, Dhruv Batra, Akshara Rai,
  and Roozbeh Mottaghi.
\newblock {Habitat 3.0: A Co-Habitat for Humans, Avatars and Robots}.
\newblock \emph{arXiv preprint arXiv:2310.13724}, 2023.

\bibitem[Reed et~al.(2022)Reed, Zolna, Parisotto, Colmenarejo, Novikov,
  Barth-maron, Gim{\'e}nez, Sulsky, Kay, Springenberg,
  et~al.]{reed2022generalist}
Scott Reed, Konrad Zolna, Emilio Parisotto, Sergio~G{\'o}mez Colmenarejo,
  Alexander Novikov, Gabriel Barth-maron, Mai Gim{\'e}nez, Yury Sulsky, Jackie
  Kay, Jost~Tobias Springenberg, et~al.
\newblock {A Generalist Agent}.
\newblock \emph{Transactions on Machine Learning Research}, 2022.

\bibitem[Sanchez et~al.(2023)Sanchez, Fan, Spangher, Levi, Ammanamanchi, and
  Biderman]{sanchez2023stay}
Guillaume Sanchez, Honglu Fan, Alexander Spangher, Elad Levi, Pawan~Sasanka
  Ammanamanchi, and Stella Biderman.
\newblock {Stay on topic with Classifier-Free Guidance}.
\newblock \emph{arXiv preprint arXiv:2306.17806}, 2023.

\bibitem[Savva et~al.(2019)Savva, Kadian, Maksymets, Zhao, Wijmans, Jain,
  Straub, Liu, Koltun, Malik, et~al.]{savva2019habitat}
Manolis Savva, Abhishek Kadian, Oleksandr Maksymets, Yili Zhao, Erik Wijmans,
  Bhavana Jain, Julian Straub, Jia Liu, Vladlen Koltun, Jitendra Malik, et~al.
\newblock {Habitat: A Platform for Embodied AI Research}.
\newblock In \emph{International Conference on Computer Vision}, 2019.

\bibitem[Shridhar et~al.(2020)Shridhar, Thomason, Gordon, Bisk, Han, Mottaghi,
  Zettlemoyer, and Fox]{shridhar2020alfred}
Mohit Shridhar, Jesse Thomason, Daniel Gordon, Yonatan Bisk, Winson Han,
  Roozbeh Mottaghi, Luke Zettlemoyer, and Dieter Fox.
\newblock {ALFRED: A Benchmark for Interpreting Grounded Instructions for
  Everyday Tasks}.
\newblock In \emph{Computer Vision and Pattern Recognition}, 2020.

\bibitem[Silver et~al.(2016)Silver, Huang, Maddison, Guez, Sifre, Van
  Den~Driessche, Schrittwieser, Antonoglou, Panneershelvam, Lanctot,
  et~al.]{silver2016mastering}
David Silver, Aja Huang, Chris~J Maddison, Arthur Guez, Laurent Sifre, George
  Van Den~Driessche, Julian Schrittwieser, Ioannis Antonoglou, Veda
  Panneershelvam, Marc Lanctot, et~al.
\newblock {Mastering the game of Go with deep neural networks and tree search}.
\newblock \emph{Nature}, 529\penalty0 (7587):\penalty0 484, 2016.

\bibitem[Silver et~al.(2018)Silver, Hubert, Schrittwieser, Antonoglou, Lai,
  Guez, Lanctot, Sifre, Kumaran, Graepel, et~al.]{silver2018general}
David Silver, Thomas Hubert, Julian Schrittwieser, Ioannis Antonoglou, Matthew
  Lai, Arthur Guez, Marc Lanctot, Laurent Sifre, Dharshan Kumaran, Thore
  Graepel, et~al.
\newblock {A general reinforcement learning algorithm that masters chess,
  shogi, and Go through self-play}.
\newblock \emph{Science}, 362\penalty0 (6419):\penalty0 1140--1144, 2018.

\bibitem[Srivastava et~al.(2021)Srivastava, Li, Lingelbach, Mart\'in-Mart\'in,
  Xia, Vainio, Lian, Gokmen, Buch, Liu, Savarese, Gweon, Wu, and
  Fei-Fei]{srivastava2021behavior}
Sanjana Srivastava, Chengshu Li, Michael Lingelbach, Roberto Mart\'in-Mart\'in,
  Fei Xia, Kent Vainio, Zheng Lian, Cem Gokmen, Shyamal Buch, Karen Liu, Silvio
  Savarese, Hyowon Gweon, Jiajun Wu, and Li~Fei-Fei.
\newblock {BEHAVIOR: Benchmark for Everyday Household Activities in Virtual,
  Interactive, and Ecological Environments}.
\newblock In \emph{Conference in Robot Learning}, 2021.

\bibitem[Stone et~al.(2023)Stone, Xiao, Lu, Gopalakrishnan, Lee, Vuong,
  Wohlhart, Zitkovich, Xia, Finn, et~al.]{stone2023open}
Austin Stone, Ted Xiao, Yao Lu, Keerthana Gopalakrishnan, Kuang-Huei Lee, Quan
  Vuong, Paul Wohlhart, Brianna Zitkovich, Fei Xia, Chelsea Finn, et~al.
\newblock {Open-World Object Manipulation using Pre-trained Vision-Language
  Models}.
\newblock \emph{arXiv preprint arXiv:2303.00905}, 2023.

\bibitem[Szot et~al.(2021)Szot, Clegg, Undersander, Wijmans, Zhao, Turner,
  Maestre, Mukadam, Chaplot, Maksymets, et~al.]{szot2022habitat}
Andrew Szot, Alexander Clegg, Eric Undersander, Erik Wijmans, Yili Zhao, John
  Turner, Noah Maestre, Mustafa Mukadam, Devendra~Singh Chaplot, Oleksandr
  Maksymets, et~al.
\newblock {Habitat 2.0: Training Home Assistants to Rearrange their Habitat}.
\newblock In \emph{Advances in Neural Information Processing Systems}, 2021.

\bibitem[Tam et~al.(2022)Tam, Rabinowitz, Lampinen, Roy, Chan, Strouse, Wang,
  Banino, and Hill]{tam2022semantic}
Allison Tam, Neil Rabinowitz, Andrew Lampinen, Nicholas~A Roy, Stephanie Chan,
  DJ~Strouse, Jane Wang, Andrea Banino, and Felix Hill.
\newblock {Semantic Exploration from Language Abstractions and Pretrained
  Representations}.
\newblock In \emph{Advances in Neural Information Processing Systems}, 2022.

\bibitem[Tan et~al.(2024)Tan, Ding, Zhang, Li, Zhou, Yue, Xia, Jiang, Zheng,
  Xu, et~al.]{tan2024towards}
Weihao Tan, Ziluo Ding, Wentao Zhang, Boyu Li, Bohan Zhou, Junpeng Yue,
  Haochong Xia, Jiechuan Jiang, Longtao Zheng, Xinrun Xu, et~al.
\newblock {Towards General Computer Control: A Multimodal Agent for Red Dead
  Redemption II as a Case Study}.
\newblock \emph{arXiv preprint arXiv:2403.03186}, 2024.

\bibitem[Tesauro et~al.(1995)]{tesauro1995temporal}
Gerald Tesauro et~al.
\newblock {Temporal Difference Learning and TD-Gammon}.
\newblock \emph{Communications of the ACM}, 38\penalty0 (3):\penalty0 58--68,
  1995.

\bibitem[Tessler et~al.(2017)Tessler, Givony, Zahavy, Mankowitz, and
  Mannor]{tessler2017deep}
Chen Tessler, Shahar Givony, Tom Zahavy, Daniel Mankowitz, and Shie Mannor.
\newblock {A Deep Hierarchical Approach to Lifelong Learning in Minecraft}.
\newblock In \emph{Proceedings of the AAAI Conference on Artificial
  Intelligence}, 2017.

\bibitem[Todorov et~al.(2012)Todorov, Erez, and Tassa]{todorov2012mujoco}
Emanuel Todorov, Tom Erez, and Yuval Tassa.
\newblock {MuJoCo: A physics engine for model-based control}.
\newblock In \emph{IEEE International Conference on Intelligent Robots and
  Systems}, 2012.

\bibitem[Vemprala et~al.(2023)Vemprala, Bonatti, Bucker, and
  Kapoor]{vemprala2023chatgpt}
Sai Vemprala, Rogerio Bonatti, Arthur Bucker, and Ashish Kapoor.
\newblock {ChatGPT for Robotics: Design Principles and Model Abilities}.
\newblock \emph{arXiv preprint arXiv:2306.17582}, 2023.

\bibitem[Villegas et~al.(2022)Villegas, Babaeizadeh, Kindermans, Moraldo,
  Zhang, Saffar, Castro, Kunze, and Erhan]{villegas2022phenaki}
Ruben Villegas, Mohammad Babaeizadeh, Pieter-Jan Kindermans, Hernan Moraldo,
  Han Zhang, Mohammad~Taghi Saffar, Santiago Castro, Julius Kunze, and Dumitru
  Erhan.
\newblock {Phenaki: Variable Length Video Generation from Open Domain Textual
  Descriptions}.
\newblock In \emph{International Conference on Learning Representations}, 2022.

\bibitem[Vinyals et~al.(2019)Vinyals, Babuschkin, Czarnecki, Mathieu, Dudzik,
  Chung, Choi, Powell, Ewalds, Georgiev, et~al.]{vinyals2019grandmaster}
Oriol Vinyals, Igor Babuschkin, Wojciech~M Czarnecki, Micha{\"e}l Mathieu,
  Andrew Dudzik, Junyoung Chung, David~H Choi, Richard Powell, Timo Ewalds,
  Petko Georgiev, et~al.
\newblock {Grandmaster level in StarCraft II using multi-agent reinforcement
  learning}.
\newblock \emph{Nature}, 575\penalty0 (7782):\penalty0 350--354, 2019.

\bibitem[Wang et~al.(2023{\natexlab{a}})Wang, Xie, Jiang, Mandlekar, Xiao, Zhu,
  Fan, and Anandkumar]{wang2023voyager}
Guanzhi Wang, Yuqi Xie, Yunfan Jiang, Ajay Mandlekar, Chaowei Xiao, Yuke Zhu,
  Linxi Fan, and Anima Anandkumar.
\newblock {Voyager: An Open-Ended Embodied Agent with Large Language Models}.
\newblock \emph{arXiv preprint arXiv:2305.16291}, 2023{\natexlab{a}}.

\bibitem[Wang et~al.(2023{\natexlab{b}})Wang, Cai, Liu, Jin, Hou, Zhang, Lin,
  He, Zheng, Yang, et~al.]{wang2023jarvis}
Zihao Wang, Shaofei Cai, Anji Liu, Yonggang Jin, Jinbing Hou, Bowei Zhang,
  Haowei Lin, Zhaofeng He, Zilong Zheng, Yaodong Yang, et~al.
\newblock {JARVIS-1: Open-World Multi-task Agents with Memory-Augmented
  Multimodal Language Models}.
\newblock \emph{arXiv preprint arXiv:2311.05997}, 2023{\natexlab{b}}.

\bibitem[Ward et~al.(2020)Ward, Bolt, Hemmings, Carter, Sanchez, Barreira,
  Noury, Anderson, Lemmon, Coe, Trochim, Handley, and Bolton]{ward2020using}
Tom Ward, Andrew Bolt, Nik Hemmings, Simon Carter, Manuel Sanchez, Ricardo
  Barreira, Seb Noury, Keith Anderson, Jay Lemmon, Jonathan Coe, Piotr Trochim,
  Tom Handley, and Adrian Bolton.
\newblock {Using Unity to Help Solve Intelligence}.
\newblock \emph{arXiv preprint arXiv:2011.09294}, 2020.

\bibitem[Yang et~al.(2023)Yang, Du, Ghasemipour, Tompson, Schuurmans, and
  Abbeel]{yang2023learning}
Mengjiao Yang, Yilun Du, Kamyar Ghasemipour, Jonathan Tompson, Dale Schuurmans,
  and Pieter Abbeel.
\newblock {Learning Interactive Real-World Simulators}.
\newblock \emph{arXiv preprint arXiv:2310.06114}, 2023.

\bibitem[Yu et~al.(2020)Yu, Quillen, He, Julian, Hausman, Finn, and
  Levine]{yu2020meta}
Tianhe Yu, Deirdre Quillen, Zhanpeng He, Ryan Julian, Karol Hausman, Chelsea
  Finn, and Sergey Levine.
\newblock {Meta-World: A Benchmark and Evaluation for Multi-Task and Meta
  Reinforcement Learning}.
\newblock In \emph{Conference on Robot Learning}, 2020.

\bibitem[Zeng et~al.(2021)Zeng, Florence, Tompson, Welker, Chien, Attarian,
  Armstrong, Krasin, Duong, Sindhwani, et~al.]{zeng2022transporter}
Andy Zeng, Pete Florence, Jonathan Tompson, Stefan Welker, Jonathan Chien,
  Maria Attarian, Travis Armstrong, Ivan Krasin, Dan Duong, Vikas Sindhwani,
  et~al.
\newblock {Transporter Networks: Rearranging the Visual World for Robotic
  Manipulation}.
\newblock In \emph{Conference on Robot Learning}, 2021.

\bibitem[Zeng et~al.(2022)Zeng, Attarian, Choromanski, Wong, Welker, Tombari,
  Purohit, Ryoo, Sindhwani, Lee, et~al.]{zeng2022socratic}
Andy Zeng, Maria Attarian, Krzysztof~Marcin Choromanski, Adrian Wong, Stefan
  Welker, Federico Tombari, Aveek Purohit, Michael~S Ryoo, Vikas Sindhwani,
  Johnny Lee, et~al.
\newblock {Socratic Models: Composing Zero-Shot Multimodal Reasoning with
  Language}.
\newblock In \emph{International Conference on Learning Representations}, 2022.

\bibitem[Zolna et~al.(2024)Zolna, Cabi, Chen, Lau, Fantacci, Pasukonis,
  Springenberg, and Colmenarejo]{zolna2024gats}
Konrad Zolna, Serkan Cabi, Yutian Chen, Eric Lau, Claudio Fantacci, Jurgis
  Pasukonis, Jost~Tobias Springenberg, and Sergio~Gomez Colmenarejo.
\newblock {GATS: Gather-Attend-Scatter}.
\newblock \emph{arXiv preprint arXiv:2401.08525}, 2024.

\end{thebibliography}

\clearpage
\twocolumn
\parindent0pt
\section*{Author contributions}
In this section, we summarize author contributions by project area, role in the area, and then alphabetically per role. A role key is provided at the end.

\textbf{Agents \& models}\\[0.5em]
\emph{Leads:}\\
Andrew Lampinen\\
Hubert Soyer

\vspace{0.1cm}
\emph{Partial Leads:}\\
Danilo J. Rezende\\
Thomas Keck\\
Alexander Lerchner\\
Tim Scholtes

\vspace{0.1cm}
\emph{Past Leads:}\\
Arun Ahuja\\
Ishita Dasgupta

\vspace{0.1cm}
\emph{Core Contributors:}\\
Jeff Clune\\
Martin Engelcke\\
Ryan Faulkner\\
Karol Gregor\\
Rosemary Ke\\
Kavya Kopparapu\\
Yulan Liu\\
Joseph Marino\\
Hamza Merzic\\
Anna Mitenkova\\
Aneesh Pappu\\
John Reid\\
Daniel P. Sawyer\\
Daniel Slater\\
Heiko Strathmann\\
Allison Tam\\
Bojan Vujatovic\\
Zhengdong Wang

\vspace{0.1cm}
\emph{Contributors:}\\
Stephanie Chan\\
Kshitij Gupta\\
Drew A. Hudson\\
Jony Hudson\\
Junkyung Kim\\
Loic Matthey\\
Pierre Harvey Richemond\\
Denis Teplyashin

\newpage

\textbf{Data}\\[0.5em]
\emph{Leads:}\\
Tayfun Terzi\\
Jane Wang

\vspace{0.1cm}
\emph{Core Contributors:}\\
Junkyung Kim\\
Oscar Knagg\\
Renke Pan

\vspace{0.1cm}
\emph{Contributors:}\\
Zhitao Gong\\
Jony Hudson\\
Andrew Lampinen\\
Anna Mitenkova\\
Yani Donchev\\
Davide Vercelli\\
John Reid

\textbf{Environments: external}\\[0.5em]
\emph{Leads:}\\
Frederic Besse\\
Tim Harley\\
Piermaria Mendolicchio

\vspace{0.1cm}
\emph{Core Contributors:}\\
Sarah Chakera\\
Vikki Copeman\\
Yani Donchev\\
Arne Olav Hallingstad\\
Maria Loks-Thompson\\
Tyson Roberts\\
Peter Stys

\vspace{0.1cm}
\emph{Contributors:}\\
Charles Gbadamosi\\
Davide Vercelli\\
Duncan Williams

\textbf{Environments: internal}\\[0.5em]
\vspace{0.1cm}
\emph{Leads:}\\
David Reichert

\vspace{0.1cm}
\emph{Past Leads:}\\
Alex Cullum

\vspace{0.1cm}
\emph{Core Contributors:}\\
Andrew Bolt\\
Bethanie Brownfield\\
Sarah Chakera\\
Dario de Cesare\\
Charles Gbadamosi\\
Mimi Jasarevic\\
Laura Kampis\\
Marjorie Limont\\
Piermaria Mendolicchio\\
Yanko Oliveira\\
Alex Platonov\\
Ollie Purkiss\\
Giles Ruscoe\\
Tasha Sandars\\
Guy Simmons\\
Nathaniel Wong\\
Nick Young

\vspace{0.1cm}
\emph{Contributors:}\\
Catarina Barros\\
Gavin Buttimore\\
Adrian Collister\\
Julia Di Trapani\\
Emma Dunleavy\\
Sam Haves\\
Rory Lawton\\
Siobhan Mcloughlin\\
Valeria Oliveira\\
Haroon Qureshi\\
Davide Vercelli\\
Marcus Wainwright\\
Sarah York

\vspace{0.1cm}
\emph{Advisors:}\\
Adrian Bolton\\
Max Cant

\textbf{Evaluation}\\[0.5em]
\emph{Leads:}\\
Laura Kampis

\vspace{0.1cm}
\emph{Partial Leads:}\\
Tim Harley\\
Andrew Lampinen

\vspace{0.1cm}
\emph{Core Contributors:}\\
Martin Engelcke\\
Loic Matthey\\
Tim Scholtes\\
Daniel Slater\\
Davide Vercelli

\vspace{0.1cm}
\emph{Contributors:}\\
Bethanie Brownfield\\
Sarah Chakera\\
Anna Mitenkova\\
David Reichert\\
John Reid\\
Jaume Sanchez Elias\\
Peter Stys\\
Jane Wang

\textbf{Partnerships \& legal}\\[0.5em]
\emph{Leads:}\\
Maria Abi Raad\\
Ed Hirst\\
Alexandre Moufarek

\emph{Core Contributors:}\\
Kathryn Martin Cussons\\
Piermaria Mendolicchio

\textbf{Project}\\[0.5em]
\emph{Concept:}\\
Frederic Besse\\
Tim Harley\\
Shane Legg

\emph{Project Leads:}\\
Frederic Besse\\
Tim Harley\\
Hannah Openshaw

\vspace{0.1cm}
\emph{Past Project Leads:}\\
Felix Hill\\
Shane Legg

\vspace{0.1cm}
\emph{Technical Leads:}\\
Thomas Keck\\
Tayfun Terzi

\vspace{0.1cm}
\emph{Core Contributors:}\\
Lucy Gonzales\\
Steph Hughes-Fitt

\vspace{0.1cm}
\emph{Product Manager:}\\
Alexandre Moufarek

\vspace{0.1cm}
\emph{Advisors:}\\
Jeff Clune\\
Daan Wierstra

\textbf{Writing \& design}\\[0.5em]
\emph{Leads:}\\
Andrew Lampinen\\
Joseph Marino

\vspace{0.1cm}
\emph{Core Contributors:}\\
Martin Engelcke\\
Tim Harley\\
Laura Kampis\\
Yulan Liu\\
Daniel P. Sawyer\\
Jane Wang\\
Zhengdong Wang

\vspace{0.1cm}
\emph{Contributors:}\\
Frederic Besse\\
Max Cant\\
Jeff Clune\\
Frankie Garcia\\
David Reichert\\

\textbf{\emph{Role key}}\\
\emph{Lead}: Responsible for the project area for the whole duration of the project.\\
\emph{Partial or Past Lead}: Responsible for the project area for a part of the project duration.\\
\emph{Core Contributor}: Contributed to the project area for an extended period of time.\\
\emph{Contributor}: Contributed to the project area for a shorter period of time.\\
\emph{Advisor}: Provided advice, feedback, and guidance to the project area.\\
\emph{Project Lead}: Responsible for all aspects of the project for the whole duration of the project.\\
\emph{Past Project Lead}: Responsible for all aspects of the project for a part of the project duration.\\
\emph{Technical Lead}: Responsible for the technical direction of the project.\\

\end{document}